\documentclass[12pt]{article}

\usepackage{amssymb}
\usepackage{amsmath}
\usepackage{booktabs}
\usepackage{graphicx}
\usepackage[colorlinks,citecolor=blue,urlcolor=blue]{hyperref}
\usepackage{longtable}
\usepackage{natbib}

\setcitestyle{authoryear,open={((},close={))}} 

\begin{document}

\begin{center}
{\Large An Explainable Pipeline for Machine Learning with Functional Data}\\
\vspace{0.25cm}
Katherine Goode\textsuperscript{a}, J. Derek Tucker\textsuperscript{a}, Daniel Ries\textsuperscript{a}, Heike Hofmann\textsuperscript{b}\\
\vspace{0.25cm}
\textsuperscript{a}\emph{Sandia National Laboratories}\\
\textsuperscript{b}\emph{Department of Statistics, University of Nebraska-Lincoln}
\end{center}

Machine learning models have shown success in applications with an
objective of prediction, but the algorithmic complexity of some models
makes them difficult to interpret. Methods have been proposed to provide
insight into these ``black-box'' models, but there is little research
that focuses on supervised machine learning when the model inputs are
functional data. In this work, we consider two applications from
high-consequence spaces (national security and forensic science) with
objectives of making predictions using functional data inputs. One
application aims to classify material types to identify explosive
materials given hyperspectral computed tomography scans of the
materials. The other application considers the forensics science task of
connecting an inkjet printed document to the source printer using color
signatures extracted by Raman spectroscopy. This task is pertinent in
cases where inkjet printers are used for illicit activities such as
counterfeit currency. An instinctive route to consider for analyzing
these data is a data driven machine learning model for classification,
but due to the high consequence nature of the applications, we argue it
is important to appropriately account for the nature of the data in the
analysis to not obscure or misrepresent patterns. As such, we propose
the
\emph{\textbf{V}ariable importance \textbf{E}xplainable \textbf{E}lastic \textbf{S}hape \textbf{A}nalysis (VEESA) pipeline}
for training machine learning models with functional data that (1)
accounts for the vertical and horizontal variability in the functional
data and (2) provides an explanation in the original data space of how
the model uses variability in the functional data for prediction. The
VEESA pipeline makes use of elastic functional principal components
analysis (efPCA) to generate uncorrelated model inputs and permutation
feature importance (PFI) to identify the principal components important
for prediction. Ultimately, visualizations are used to depict the
variability captured by the important principal components. The visuals
are represented in the original data space to help subject matter
experts make informed decisions. We additionally discuss ideas for
natural extensions of the VEESA pipeline and challenges for future
research.

\vspace{0.5cm}


\section{\texorpdfstring{Introduction
\label{intro}}{Introduction }}\label{introduction}

Many machine learning models are considered ``black-boxes'' since their
algorithmic complexity results in the inability to assign a physical
meaning to the model parameters. In supervised learning, the
interpretation of model parameters provides an understanding of the
relationships between predictor and response variables. Understanding
these relationships is beneficial for multiple reasons such as model
assessment (e.g., is the model using variables in a scientifically
reasonable manner?), data insight (e.g., what patterns in the data did
the model find useful for prediction?), and helping to build trust in a
prediction (e.g., do we feel comfortable making a decision based on how
the model made use of the data to produce a prediction?). In lower
consequence applications (e.g., movie recommendations or personalized
advertisements), a lack of model ``interpretability'' may be permissible
since incorrect predictions have minimal negative impacts, but in
high-consequence areas such as national security and forensics science,
it is difficult to motivate the use of a non-interpretable model where
model transparency is essential to avoid serious mistakes. Regardless of
the consequence level of the application, interpretability provides
information that helps gain data insight and improve models.

The desire to contextually explain how predictor variables relate to
black-box predictions has led to a recent explosion of research focused
on providing insight into black-box models
\citep{adadi:2018, guidotti:2018, zhang:2020}. This area of research is
often referred to as explainable/interpretable machine learning
(EML/IML) \citep{rudin:2019}/\citep{molnar:2020a} or interpretable
artificial intelligence (XAI) \citep{adadi:2018}. While previous
research considers various model and data types, there is minimal
literature on EML for supervised learning with functional data inputs.

Each observation in a functional dataset consists of a collection of
points representing a continuous curve or surface over a compact domain
(e.g., a fixed length of time or region of space). Examples of
functional data include the heights of children measured over time
\citep{ramsay:2005} and silhouettes of animals extracted from images
\citep{srivastava:2016}. Modern technology has made the collection of
functional data commonplace with devices such as glucose monitors
\citep{danne:2017}, fitness trackers \citep{henriksen:2018}, and
environmental sensors \citep{butts:2020}. Functional data provide
detailed information of a continuous process, but the form of the data
requires special consideration to appropriately account for the
functional structure.

One approach to analyzing functional data is to compute relevant summary
statistics of the functions. This approach allows for the use of simpler
statistical analyses but may result in loss of information and incorrect
inference \citep{tucker:2013, srivastava:2016}. Instead, methods have
been developed that treat the underlying curve as an infinite
dimensional continuous function such as functional regression,
functional ANOVA, and functional principal component analysis (fPCA)
\citep{ramsay:2005}.

In machine learning, functional data are used as inputs to predict an
outcome based on the functional shapes (e.g., \citet{li:2014},
\citet{zhang:2019}, and \citet{ries:2023}). Various modeling approaches
have been suggested (e.g., \citet{rossi:2005}, \citet{tian:2010}, and
\citet{thind:2023}). One approach is to create vectors of values across
functions at each domain location (e.g., time or location) and use these
\emph{``cross-sectional''} vectors as model input features
\citep{tian:2010}. A clear downside to the cross-sectional approach is
that the dependence between observations within a function is ignored.
Another approach is to apply a dimension reduction technique, such as
fPCA, and use a subset of the transformed features as predictor
variables \citep{rossi:2005, tian:2010}.

A few papers consider explainability with functional data inputs.
\citet{martin:2014} and \citet{thind:2023} adapt the methodology of
support vector machines and neural networks, respectively, to account
for functional data. The models are constructed so the parameters can be
visualized and interpreted in the context of the functional data domain.
\citet{goode:2020} use fPCA for dimension reduction and feature
importance to identify important functional principal components (fPCs).

All previously proposed methods for using functional data as inputs to
machine learning models only account for \emph{vertical} functional
variability, where vertical variability (also known as \(y\) or
amplitude variability) is the variability in the height of functions.
\citet{tucker:2013} highlights that functional data possess both
vertical and \emph{horizontal} variability. Horizontal variability (also
known as \(x\) or phase variability) is the variability in the location
of peaks and valleys of the functions. (See \citet{tucker:2013} for
mathematical definitions.) If one variability is ignored, the resulting
functional data analysis may not accurately capture the structure in the
data.

In this paper, we consider two classification tasks with functional data
from high consequence decision spaces. The first application aims to
identify explosive materials given hyperspectral computed tomography
scans of the material for national security purposes. The second
application is motivated by illicit uses of inkjet printers, such as the
creation of counterfeit currency. Forensic investigators are interested
in being able to identify the source printer for a printer document
based on Raman spectroscopy signatures extracted from the document. We
propose a novel explainable machine learning pipeline for functional
data inputs to analyze these data. The approach, which we refer to as
the
\emph{\textbf{V}ariable importance \textbf{E}xplainable \textbf{E}lastic \textbf{S}hape \textbf{A}nalysis (VEESA) pipeline},
(1) accounts for the vertical and horizontal variability in the
functional data and (2) provides an explanation in the original data
space of how the model uses variability in the functional data for
prediction.

The pipeline uses a principal component analysis for functional data
derived using the elastic shape analysis (ESA) framework
\citep{joshi:2007, srivastava:2011, tucker:2013} referred to as
\emph{elastic fPCA (efPCA)}. Similar to the use of fPCA with machine
learning, the observed functions are converted to fPCs and used as
predictor variables, but unlike fPCA, efPCA accounts for vertical and
horizontal functional variability. Elastic functional principal
components (efPCs) have been used previously in modeling approaches
\citep{tucker:2019, tucker:2020b}, but there no literature on the use of
efPCA with explainability for machine learning. After the model is
trained, the model agnostic method of permutation feature importance
(PFI) \citep{fisher:2019} is applied to identify important efPCs. PFI is
known to be biased when correlation is present between variables
\citep{hooker:2021}, but since principal components (PCs) are
uncorrelated, there is no concern of feature importance bias due to
correlation in the VEESA pipeline. Visualization is used to interpret
the important efPCs. The visualizations depict the variability captured
by the important efPCs in the original data space, which allows both
model developers and subject matter experts to determine whether the
model is drawing on reasonable phenomenological characteristics for
prediction.

This paper is structured as follows. In Section \ref{background},
background is provided on the methods of efPCA and PFI. Section
\ref{methods} describes the VEESA pipeline in detail. Section
\ref{examples} presents the VEESA pipeline analyses of the explosive
material and inkjet printer applications. We discuss the advantages and
limitations of the VEESA pipeline and possible future research
directions in Section \ref{discussion}.

\section{\texorpdfstring{Background
\label{background}}{Background }}\label{background}

In this section, we provide background on the separation of vertical and
horizontal functional variability (Section \ref{separation}), efPCA
(Section \ref{efpca}), and PFI (Section \ref{pfi}). We introduce a
simulated dataset based on an example from \citet{tucker:2013} to
demonstrate these methods. We will refer to this data as the
\emph{shifted peaks} data. The shifted peaks data are available from the
\emph{veesa} R package \citep{goode:2024}, and the code associated with
all analyses of the shifted peaks data in this paper is available at
\url{github.com/sandialabs/veesa/tree/master/demos/goode-et-al-paper}.

\begin{figure}

{\centering \includegraphics[width=5.5in]{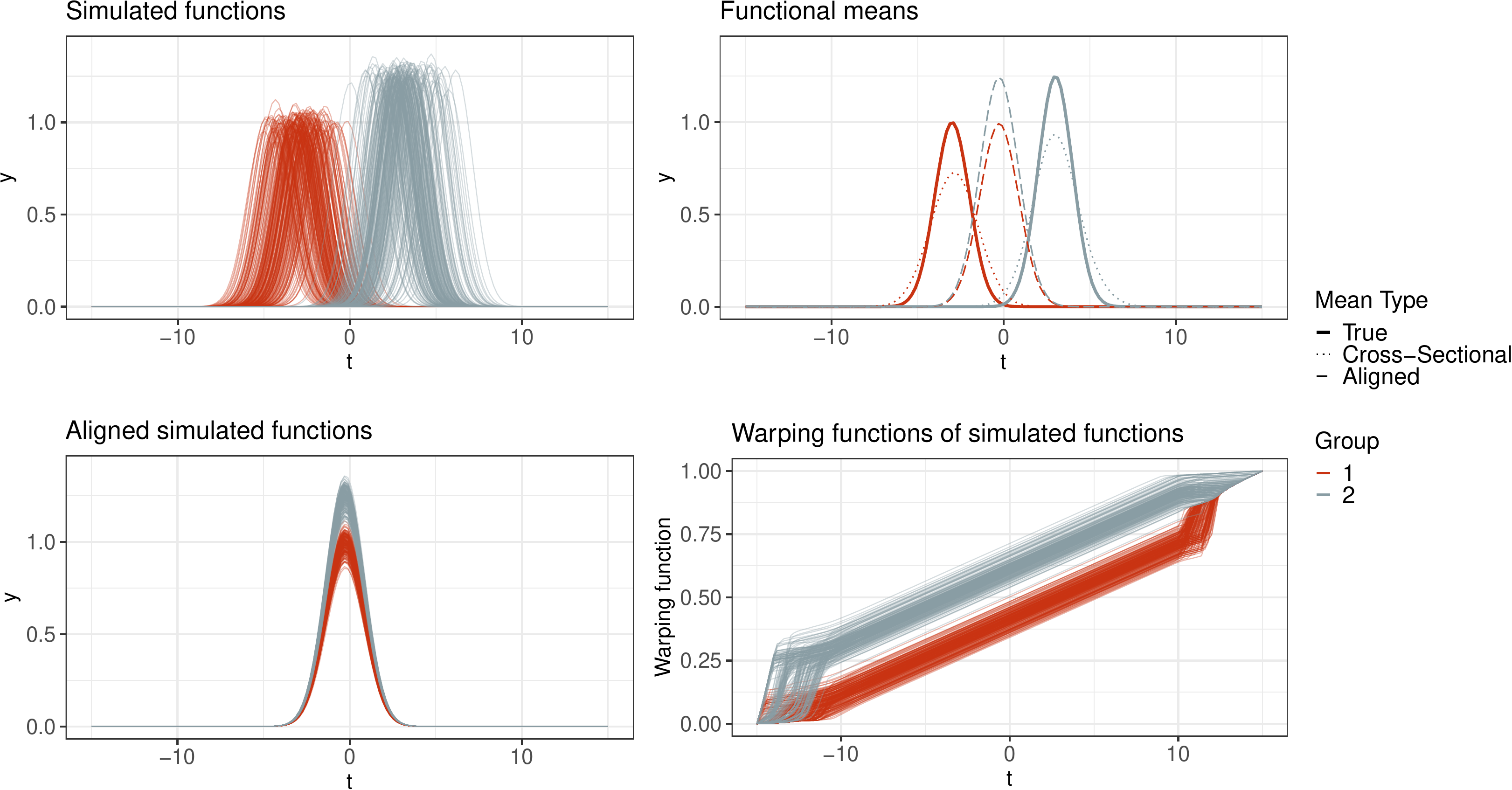} 

}

\caption{(Top Left) Training data functions from the shifted peak simulated data. (Top Right) The true, cross-sectional, and aligned functional means. (Bottom Left) The aligned functions. (Bottom Right) The warping functions from the alignment of the functions.}\label{fig:fig1}
\end{figure}

The shifted peaks data contain 500 functions from two groups with 250
functions per group (Figure \ref{fig:fig1} (top left)). The functions
are simulated as

\begin{equation}
y_{g,i}(t) = z_{g,i}e^{-(t-a_{g,i})^2/2} \label{eq:sim}
\end{equation}

\noindent where \(g=1,2\) indicates the group, \(i=1,2,...,\) 250
identifies a function within a group, \(t\in\) {[}-15, 15{]},
\(z_{g,i}\overset{iid}{\sim}N(\mu_{z,g},(0.05)^2)\), and
\(a_{g,i}\overset{iid}{\sim}N(\mu_{a,g},(1.25)^2)\). We set
\(\mu_{z,1}=\) 1, \(\mu_{z,2}=\) 1.25, \(\mu_{a,1}=\) -3, and
\(\mu_{a,2}=\) 3, so the true functional mean of group 1 has a peak that
is lower and further to the left than group 2 (solid lines in Figure
\ref{fig:fig1} (top right)). The functions are generated with 150
equally spaced observations per function. For prediction, the functions
are randomly separated into training and testing sets with 400 and 100
functions, respectively.

\subsection{\texorpdfstring{Separation of Vertical and Horizontal
Variability
\label{separation}}{Separation of Vertical and Horizontal Variability }}\label{separation-of-vertical-and-horizontal-variability}

As previously mentioned, there are two types of variability present in
functional data: vertical and horizontal variability. The shifted peaks
data in Figure \ref{fig:fig1} (top left) exemplify vertical variability
by the differences in peak intensity and horizontal variability by the
difference in peak times. The dotted lines in Figure \ref{fig:fig1} (top
right) demonstrate an example of how ignoring horizontal variability can
lead to inaccurate representations of the functional forms. The dotted
lines represent the cross-sectional group means of the shifted peaks
data (i.e., mean value computed across all functions at a time). Both
cross-sectional means have different shapes than the true means, which
are narrower and have higher peaks than the cross-sectional means.

\citet{tucker:2013} account for both variabilities in efPCA by first
decomposing observed functions into two new sets of functions: aligned
and warping functions. Aligned functions match the peaks and valleys
from the observed functions, and thus, capture the vertical variability
present in the observed functions. Warping functions transform the
observed functions to the aligned functions, and thus, capture the
horizontal variability in the observed functions. The separation process
described in \citet{tucker:2013} uses methodology from the ESA framework
described in \citet{srivastava:2011} and \citet{joshi:2007}. Here, we
provide an overview of the separation process and refer the reader to
those papers and \citet{srivastava:2016} for more information.

We start by letting \(f\) be a real-valued function with the domain
\([0,1]\); this domain can be easily generalized to any other compact
subinterval of \(\mathbb{R}^1\). We assume that all functions considered
in the following analysis are observed on the same interval.
Furthermore, for concreteness, only functions that are absolutely
continuous on \([0,1]\) will be considered, and we let \(\mathcal{F}\)
denote the set of all such functions. In practice, the observed data are
discrete, so this assumption is not a restriction. Also, let \(\Gamma\)
be the set of orientation-preserving diffeomorphisms of the unit
interval \([0,1]\):
\[\Gamma = \{\gamma: [0,1] \rightarrow [0,1] |~\gamma(0) = 0,~\gamma(1)=1,\gamma~\textnormal{is a diffeomorphism} \}.\]
Elements of \(\Gamma\) play the role of warping functions. For any
\(f \in \mathcal{F}\) and \(\gamma \in \Gamma\), the composition
\(f \circ \gamma\) denotes the time warping of \(f\) by \(\gamma\)
(i.e., the aligned version of \(f\)).

As described in \cite{tucker:2013}, there are two metrics to measure the
amplitude and phase variability of functions. These metrics are proper
distances, one on the quotient space \(\mathcal{F}/\Gamma\) (i.e.,
amplitude) and the other on the group \(\Gamma\) (i.e., phase). The
amplitude or \(y\)-distance for any two functions
\(f_1,\ f_2 \in \mathcal{F}\) is defined as \begin{equation}
d_a(f_1, f_2) = \inf_{\gamma \in \Gamma} \|q_1 - (q_2 \circ \gamma)\sqrt{\dot{\gamma}}\|,
\label{eq:d_a}
\end{equation} where
\(q(t) = \mbox{sign}(\dot{f}(t)) \sqrt{ |\dot{f}(t)|}\) is known as the
\emph{square-root velocity function (SRVF)} (\(\dot{f}\) is the time
derivative of \(f\)). The SRVF transformation of \(f\) is used in the
distance computation since if \(f\) is absolutely continuous, \(q\) is
square-integrable, which allows \(d_a(f_1,f_2)\) to be a proper
distance. This is unlike other functional alignment processes (of the
type \(\inf_{\gamma\in\Gamma}\lVert f_1-f_2\circ\gamma\rVert\)) that are
not proper distances. For more details on the SRVF transformation, see
\citet{srivastava:2011}. The distance in Equation \ref{eq:d_a} is solved
using the dynamic programming algorithm \citep{bertsekas:1996}.

The second metric is the phase or \(x\)-distance for measuring the
distance between two warping functions \(\gamma_1,\gamma_2\in\Gamma\).
Since \(\Gamma\) is an infinite-dimensional nonlinear manifold (i.e.,
not a standard Hilbert space), it is challenging to compute a distance
on \(\Gamma\). A transformation is applied to the warping functions to
simplify the complicated geometry of \(\Gamma\). Specifically, the SRVF
of \(\gamma\) is computed: \(\psi=\sqrt{\dot{\gamma}}\). Note that
\(\dot{\gamma}>0\). Let \(\Psi\) be the set of all such \(\psi\)'s,
which can be shown to be the Hilbert sphere (i.e.,
\(\Psi=\mathbb{S}^+_\infty\), the positive orthant of the Hilbert
sphere). For two warping functions \(\gamma_1,\gamma_2\in\Gamma\), the
\(x\)-distance is computed as \begin{equation}
d_p(\gamma_1, \gamma_2) = d_{\psi}(\psi_1, \psi_2)=\cos^{-1}\left(\int_0^1\psi_1(t)\psi_2(t)dt\right),
\label{eq:d_p}
\end{equation} the arc-length between the SRVFs of the warping functions
on the Hilbert sphere. Note that \(\psi\) is intervtible, which makes it
possible to reconstruct \(\gamma\) from \(\psi\) as
\(\gamma(t)=\int_0^t\psi(s)^2ds\) since \(\gamma(0)=0\).

For separating the phase-amplitude components of a set of functions
\(f_1,f_2,...,f_n\), we first compute a Karcher mean of the given
functions (denoted by \(\mu_f\in\mathcal{F}\) and \(\mu_q\) in SRVF
space) under the metric \(d_a\): \begin{eqnarray}
(\mbox{In $\mathcal{F}$ space}): \mu_f &=& \mathop{\mathrm{arg\,min}}_{f \in \mathcal{F}} \sum_{i=1}^n d_a(f, f_i)^2\ \ \\
(\mbox{In SRVF space}): \mu_q &=&  \mathop{\mathrm{arg\,min}}_{q \in \mathbb{L}^2} \sum_{i=1}^n \left( \inf_{\gamma_i \in \Gamma} \| q - (q_i, \gamma_i) \|^2 \right)\ , \label{eq:Karcher-min}
\end{eqnarray} where
\((q_i,\gamma_i)=(q_i\circ\gamma_i)\sqrt{\dot{\gamma}_i}\). Note that
these formulations are equivalent. That is,
\(\mu_q=\mbox{sign}(\dot{\mu}_f)\sqrt{|\dot{\mu}_f|}\). As described in
\cite{srivastava:2011}, the algorithm for computing the Karcher mean
also results in (1) aligned functions
\(\left\{\tilde{f}_1, \tilde{f}_2,...,\tilde{f}_n\right\}\) where
\(\tilde{f}_i = f_i \circ \gamma_i\), representing the amplitude
variability and (2) the warping functions
\(\{\gamma_1, \gamma_2,..., \gamma_n\}\) used in aligning the original
data and representing the phase variability.

We apply the separation method to the shifted peaks data. Figure
\ref{fig:fig1} (bottom left and bottom right) shows the aligned and
warping functions, respectively. The dashed lines in Figure
\ref{fig:fig1} (top right) are the cross-sectional group means from the
aligned functions. With the horizontal variability removed, the group
means of the aligned functions correctly capture the shape of the true
group means (ignoring peak timing).

\subsection{\texorpdfstring{Elastic Functional Principal Component
Analysis
\label{efpca}}{Elastic Functional Principal Component Analysis }}\label{elastic-functional-principal-component-analysis}

There are three efPCA methods that model functional variability:
\emph{vertical, horizontal, and joint fPCA (vfPCA, hfPCA, and jfPCA)}.
vfPCA and hfPCA are proposed in \citet{tucker:2013}. As their names
suggest, vfPCA and hfPCA provide separate evaluations of the horizontal
and vertical variabilities present in the functions using the aligned
and warping functions, respectively. \citet{lee:2017a} proposed jfPCA
(or combined fPCA), which jointly accounts for vertical and horizontal
variability by concatenating the warping and aligned functions into a
combined function, \(g^C\), before computing principal components.
\citet{tucker:2020b} make a modification to the methodology proposed by
\citet{lee:2017a} by constructing the combined function \(g^C\) using
the SRVF \(\tilde{q}\) of the aligned function \(\tilde{f}\), since
\(\tilde{q}\) is guaranteed to be an element of \(\mathbb{L}^2\). We use
the modified version by \citet{tucker:2020b} in this paper. We provide a
generalized version of the process for computing efPCs here, and we
refer the reader to \citep{tucker:2013}, \citep{lee:2017a}, and
\citep{tucker:2020b} for further details.

\begin{table}[!h]

\caption{\label{tab:domains}Functional Principal Component Domains.}
\centering
\begin{tabular}[t]{lccc}
\toprule
  & Vertical fPCA & Horizontal fPCA & Joint fPCA\\
\midrule
Representation & $\tilde{q}$ & $\gamma$ & $g^C = [\tilde{q}~~Cv(t)]$\\
Variability & Amplitude & Phase & Amplitude + Phase\\
Metric & Fisher-Rao & Fisher-Rao & Fisher-Rao\\
\bottomrule
\end{tabular}
\end{table}

For a set of functions \(\{f_1,f_2,...,f_n\}\), separate the functions
into aligned
\(\left\{\tilde{f}_1, \tilde{f}_2,..., \tilde{f}_n\right\}\) and warping
\(\{\gamma_1,\gamma_2,...,\gamma_n\}\) functions using the method
described in Section \ref{separation}. Next, compute a sample covariance
function on the functional representation shown in Table
\ref{tab:domains} based on the variability of interest. For joint fPCA,
\(C\) is a constant such that \(C>0\), and \(v(t)\) is a tangent space
representation of \(\psi(t)\), which is used for computational ease. See
\citep{lee:2017a} for a data-driven approach for estimating \(C\) and
\citet{tucker:2020b} for details on the computation of \(v(t)\). For a
generalization, let \(z_1,z_2,...,z_n\) represent the set of functions
capturing the specified type of variability (i.e., \(\tilde{q}\),
\(\gamma\), or \(g^C = [\tilde{q}~~Cv(t)]\)). The sample covariance
function is computed as \begin{equation}
K=\frac{1}{n-1}\sum_{i=1}^n\left(z_i-\hat{\mu}_z\right)\left(z_i-\hat{\mu}_z\right)^T.
\label{eq:cov}
\end{equation} where \(\hat{\mu}_z\) is the sample mean function. Next,
apply singular value decomposition (SVD) to the covariance matrix to
obtain \(K=U_K\Sigma_KV_K^T\) where \(U_K\) contains the directions of
principal variability. The principal coefficients are computed as
\(\langle z_i, U_{K,j}\rangle\).

The principal directions can be visualized in the original function
space, \(\mathcal{F}\), to interpret functional variability captured by
fPCs. A common visualization is to plot the Karcher mean and the
variation in the principal directions. In an abuse of notation, we
represent a generalized form of this as \begin{equation}
\mu_z\pm\sqrt{\tau\Sigma_{K,jj}}U_{K,j}
\label{eq:prdirs}
\end{equation} where \(\tau\in\mathbb{N}^+\). By plotting a series of
curves given various values of \(\tau\), the visualization provides a
visual spectrum of the variability of shapes of functions captured by
the principal component. We refer to these as plots of the
\emph{principal direction plots}.

\begin{figure}

{\centering \includegraphics[width=4in]{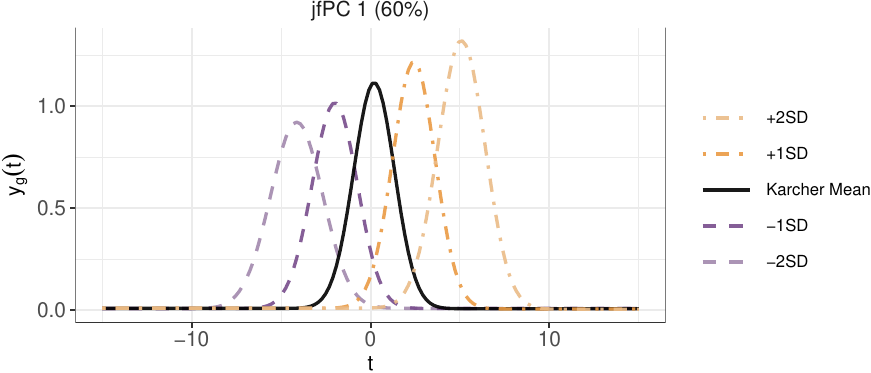} 

}

\caption{Plot of the principal directions for interpreting the functional variability captured by jfPC 1 from the shifted peaks data shown in Figure \ref{fig:fig1}.}\label{fig:fig2}
\end{figure}

Figure \ref{fig:fig2} shows the principal direction plots for the first
joint functional principal component (jfPC) from an application of jfPCA
to the shifted peaks data. The application of jfPCA is done using the
\emph{fdasrsf} R package \citep{tucker:2024r}. The solid black line
represents the Karcher functional mean. The dashed/dashed-dotted lines
represent the principal directions plus/minus 1 and 2 standard
deviations. The visual indicates that the first principal component
captures a large amount of variability in peak time and variability in
peak intensity, which agrees with the functional variability seen in
Figure \ref{fig:fig1} (top left).

\subsection{\texorpdfstring{Permutation Feature Importance
\label{pfi}}{Permutation Feature Importance }}\label{permutation-feature-importance}

Permutation feature importance (PFI) was initially developed as a tool
for random forests by \citet{breiman:2001}. \citet{fisher:2019}
recognized that the process could be generalized to any model type,
which allows for comparison of feature importance across model types.
The objective of PFI is to quantify the importance of a predictor
variable by measuring the change in model performance on a set of data
when the predictor variable is randomly permuted. The variable is
considered important if the model performance worsens considerably when
the variable is permuted in comparison to the model predictions on the
observed data.

There is some variability in how PFI has been previously defined. In
this paper, we define PFI using the following definition. Let
\(\mathcal{A}\) be a model, \(\textbf{X}\) be an \(n\times p\) data
matrix with \(n\) observations and \(p\) predictor variables, \(X_1\),
\(X_2\),\ldots, and \(X_p\) be the columns of \(\textbf{X}\), and
\(\mathfrak{m}\) be a performance metric computed on \(\textbf{X}\) with
model \(\mathcal{A}\). The data \(\textbf{X}\) may be training data,
testing data, or another set of data of interest. PFI provides insight
into how the model \(\mathcal{A}\) makes use of the data to provide
predictions. The results may differ between the training and testing
data. The performance metric \(\mathfrak{m}\) may be any metric of
interest such that larger values indicate better performance. Examples
include negative mean squared error for regression and accuracy for
classification.

Define PFI for variable \(j\in\{1,...,p\}\) as follows. For repetition
\(r\in\{1,...,R\}\):

\begin{enumerate}
\setlength{\itemindent}{1em}
\item Randomly permute $X_j$. Define the permuted variable as $\tilde{X}_{j,r}$.
\item Create a new dataset, $\tilde{\textbf{X}}_{j,r}$, by replacing $X_j$ in $\textbf{X}$ with $\tilde{X}_{j,r}$.
\item Compute $\mathfrak{m}_{j,r}$ as the performance metric on $\tilde{\textbf{X}}_{j,r}$ for model $\mathcal{A}$.
\item Then, the PFI for variable $j$ is equal to $$\mathcal{I}_j=\mathfrak{m}-\frac{1}{R}\sum_{r=1}^K\mathfrak{m}_{j,r}.  \label{eq:pfi}$$
\end{enumerate}

\noindent \(\mathcal{I}_j\) is interpreted as the average change in
model performance when \(j\) is randomly permuted. For example, if
accuracy is used for \(\mathfrak{m}\), PFI is interpreted as, ``on
average, the accuracy of model \(\mathcal{A}\) decreased by
\(\mathcal{I}_j\) when feature \(j\) is randomly permuted''. An example
of PFI computed on the shifted peaks data is included in Section
\ref{methods}.

In addition to considering the quantity of \(\mathcal{I}_j\), it may be
valuable to consider the variability of the change in model performance
across repetitions. Thus, define the feature importance for variable
\(j\) and a single repetition \(r\) as
\(\mathcal{I}_{j,r}=\mathfrak{m}-\mathfrak{m}_{j,r}\). We may compute
the standard deviation and create visualizations capturing the
variability in \(\mathcal{I}_{j,r}\) across the \(R\) repetitions. A
large variation in \(\mathcal{I}_{j,r}=\mathfrak{m}-\mathfrak{m}_{j,r}\)
indicates that the change in model performance is dependent on the
permutation of variable \(j\), and more replications may be needed to
obtain a good estimate of the average PFI.

\section{\texorpdfstring{Methods
\label{methods}}{Methods }}\label{methods}

In this section, we present the VEESA pipeline. As previously stated,
the VEESA pipeline is a procedure for training and explaining machine
learning models in applications where functional data are used to
predict an outcome. By making use of the ESA framework, the VEESA
pipeline mathematically captures the continuous nature of the observed
functions and accounts for the vertical and horizontal variability that
may be present in the functions. We first describe the application of
the VEESA pipeline to training data for model building in Section
\ref{pipeline-train}. We then describe how the pipeline is applied to
test/validation data in Section \ref{pipeline-test}. Lastly, in Section
\ref{purpose}, we describe the intended uses for the VEESA pipeline
explanations. A comparison of the results to the cross-sectional
approach is included in the supplemental material. The methodology
described here is implemented in the \emph{veesa} R package
\citep{goode:2024} and Python functions provided at
\url{github.com/sandialabs/veesa/tree/master/demos/goode-et-al-paper}.
The R and Python code use the \emph{fdasrsf} R package
\citep{tucker:2024r} and the \emph{fdasrvf} \citep{tucker:2024py}
packages, respectively, to apply functional alignment and efPCA.

\subsection{\texorpdfstring{VEESA Pipeline (Training Data)
\label{pipeline-train}}{VEESA Pipeline (Training Data) }}\label{veesa-pipeline-training-data}

\begin{figure}

{\centering \includegraphics[width=4.25in]{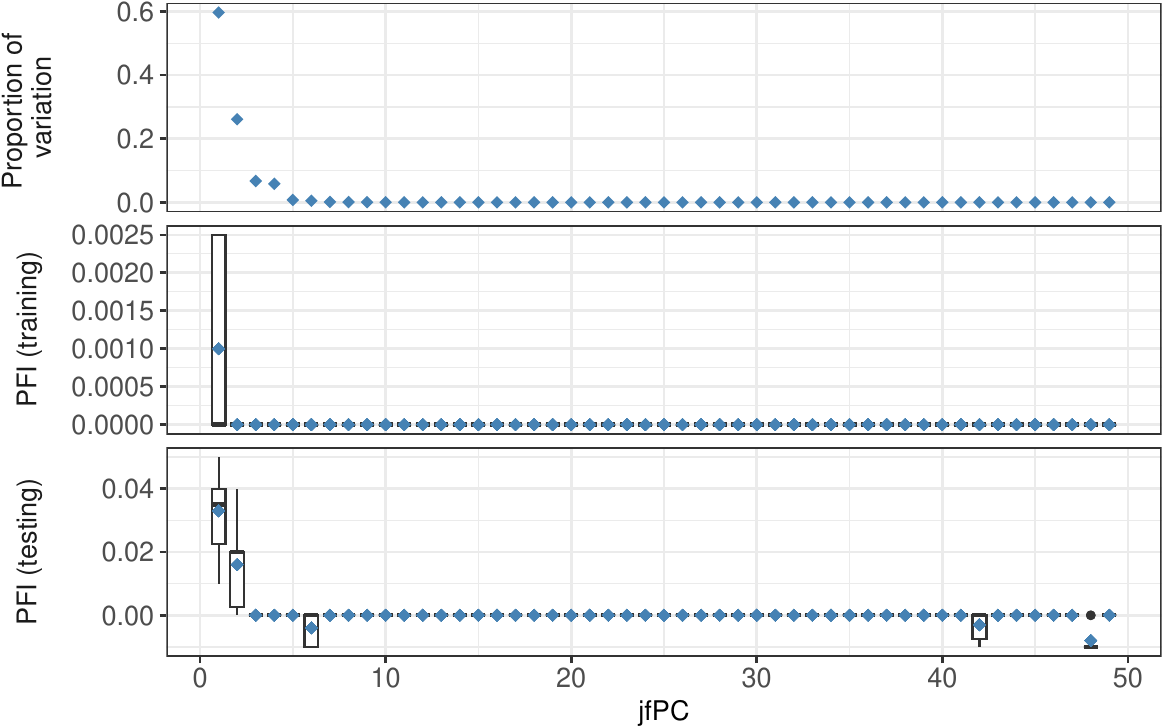} 

}

\caption{(Top) Proportion of variation explained by the jfPCs computed from the shifted peaks data. (Middle) Boxplots of PFI values across replicates associated with each jfPC from the shifted peaks data random forest computed on the training data. The blue diamonds represent the PFI value averaged over replicates. (Bottom) Same as middle plot but computed on the shifted peaks test data.}\label{fig:fig3}
\end{figure}

In order to obtain an accurate estimate of model predictive performance
on new data, supervised machine learning models are typically fit to a
set of training data and then used to obtain predictions on a set of
held out data (i.e., either testing or validation data). We begin by
describing the steps in the VEESA pipeline as applied to training data.
We include an example of each step applied to the shifted peaks data
from Section \ref{background} except for the smoothing step since the
data are simulated as smooth functions.

\subsubsection{Step 1: Smoothing}\label{step-1-smoothing}

If the observed functions are not smooth, begin by applying a smoothing
technique to the individual functions in the training data such as a box
filter \citep{tucker:2013} or splines \citep{ullah:2013}. The SRVF
transformation in the separation of vertical and horizontal variability
requires the computation of a derivative. As a result, it is recommended
that smoothing is applied to observed functions before the variability
separation process to compute an accurate estimate of the derivative,
and in turn, the SRVF. The choice of the smoothing method is left to the
analysts. In order to determine how much to smooth the functions, we
suggest treating the amount of smoothing as a tuning parameter in the
VEESA pipeline and conducting a cross-validation analysis to identify
the amount of smoothing that results in the best predictive performance.

\subsubsection{Step 2: Separation of Functional
Variability}\label{step-2-separation-of-functional-variability}

Apply the ESA alignment process described in Section \ref{separation} to
the training data. As previously described, the aligned and warping
functions from the shifted peaks training data are shown in Figure
\ref{fig:fig1}.

\subsubsection{Step 3: Elastic Functional Principal Component
Analysis}\label{step-3-elastic-functional-principal-component-analysis}

Apply one of the efPCA methods described in Section \ref{efpca} to the
training data. The selection of a method will depend on the relationship
between the functional form and the response variable. If predictive
information is only contained in the vertical or horizontal direction,
then vfPCA or hfPCA, respectively, would be applicable. However, if
predictive information is contained in both types of variability, then
jfPCA should be used. If it is not known whether predictive information
is associated with vertical variability, horizontal variability, or
both, all methods could be applied to determine which leads to the best
predictive performance. Figure \ref{fig:fig3} (top) shows the proportion
of variance explained by jfPCs computed on the shifted peaks training
data. Since we know predictive information is contained in both vertical
and horizontal variability in the shifted peaks data, we use jfPCA.

\subsubsection{Step 4: Model Training}\label{step-4-model-training}

Train a model using the efPCs obtained in Step 3 as predictor variables.
Since the amount of variability explained by a principal component does
not indicate the predictive ability of a principal component, we
recommend initially training a model with all (or a large number of)
efPCs. Feature selection may be performed after obtaining PFI results.
For the shifted peaks data, a random forest is trained using the
\emph{randomForest} package in R \citep{liaw:2022} with the default
settings. The jfPCs obtained in Step 3 are used as the predictor
variables, and the group as the response variable. The model returns an
accuracy of 1 on training data. The test data accuracy is provided in
Section \ref{pipeline-test}.

\subsubsection{Step 5: Permutation Feature
Importance}\label{step-5-permutation-feature-importance}

Apply PFI, as described in Section \ref{pfi}, to determine the
importance of the efPCs. Since efPCs are orthogonal, there is no concern
of bias in the results due to correlation between predictor variables.
The use of PFI as the explainability method provides two key advantages.
First, PFI is model agnostic, so variable importance is comparable
across model types and can potentially assist with model selection.
Second, PFI is relatively simple to understand, which makes it possible
to explain to decision makers who may have little familiarity with
machine learning models. However, other explainability methods could
easily be substituted such as partial dependence plots
\citep{friedman:2001}. The PFI values for the random forest trained on
the shifted peaks data are computed on the training data using a metric
of accuracy and 10 replications. Figure \ref{fig:fig3} (middle) shows
the PFI results for the random forest from Step 4. The boxplots depict
the PFI values across replicates, and the diamonds indicate the PFI
value averaged over replicates. jfPC 1 is clearly the most important
predictor variable in the random forest (with some replicates returning
a change in accuracy of 0). The value of PFI (averaged across
replicates) for jfPC 1 is interpreted as follows: on average, the random
forest accuracy (on the training data) decreases by 0.001 when jfPC 1 is
randomly permuted. The variability across replicates for jfPC 1 and the
relatively small change in accuracy provides some evidence that while
the PFI values are 0 for all other jfPCs, the model is drawing on
information from at least one of the other jfPCs.

\subsubsection{Step 6: efPC
Visualizations}\label{step-6-efpc-visualizations}

PFI identifies the efPCs important to the model, but visualizations are
needed to interpret the functional variability used by the model for
prediction. Specifically, the efPCs are interpreted using principal
direction plots as discussed in Section \ref{efpca}. The principal
direction plots depict the functional variability captured by a
principal component on the original data space. This representation is
advantageous to subject matter experts who are used to working with
functions on this space and likely have an intuition as to the
functional variability that would be expected to be related to a
distinction between classes. The principal direction plot for jfPC 1
from the shifted peak data (Figure \ref{fig:fig2}) indicates that jfPC 1
captures variability in both peak time and intensity based on the known
distinction between the functional group means.

\subsection{\texorpdfstring{VEESA Pipeline (Testing Data)
\label{pipeline-test}}{VEESA Pipeline (Testing Data) }}\label{veesa-pipeline-testing-data}

After the application of the VEESA pipeline to the training data, the
pipeline steps may be applied to additional data such as test data. The
general process remains the same but with some implementation
adjustments for certain steps.

\begin{enumerate}
\setlength{\itemindent}{1em}
\item \emph{Smoothing} Apply the same smoothing process as is applied to the training data.
\item \emph{Separation of Functional Variability} The alignment of test data is implemented by aligning the test data functions to the Karcher Mean of the training data SRVFs. That is, for a set of training data functions $\{f_1,f_2,...,f_n\}$ and a set of test data functions $\left\{f_{n+1}, f_{n+2},...f_{n+m}\right\}$, compute SRVFs of the test data functions $\left\{q_{n+1},q_{n+2},...,q_{n+m}\right\}$. Let $\hat{\mu}_q$ represent the sample Karcher mean of the training data functions in SRVF space. Compute the warping functions $\{\gamma_{n+1},\gamma_{n+2},...,\gamma_{m+n}\}$ that align the test data SRVFs to $\hat{\mu}_q$. That is,
\begin{equation}
\gamma_{n+j}=\underset{\gamma\in\Gamma}{\arg\min}\|\hat{\mu}_q-(q_{n+j} \circ \gamma)\sqrt{\dot{\gamma}}\|,
\label{eq:testalign}
\end{equation}
where $j=1,...,m$. The warping functions computed on the test data, $\gamma_{n+j}$, are then used to compute the aligned test data functions: $(q_{n+j} \circ \gamma)\sqrt{\dot{\gamma}_{n+j}}$ (in SRVF space).
\item \emph{Elastic Functional Principal Component Analysis} As described in Section \ref{efpca}, the specifics of computing efPCs on the test data will vary based on the efPCA method applied to the training data, but the general concept is the same. Let $z_{n+1},z_{n+2},...,z_{n+m}$ represent the set of functions computed in the alignment process of the test data associated with the desired efPCA method. Then the principal coefficients are computed as $\langle z_{n+j}, U_{K,j}\rangle$ for $j=1,...,m$, where $U_{K,j}$ contains the directions of principal variability computed when efPCA is applied to the training data. Note that this multiplication is done on the SRVF space and can then be converted back to the original space.
\item \emph{Model} The efPCs computed on the test data are input into the previously trained model to obtain predictions.
\item \emph{Permutation Feature Importance} PFI is computed on the test data. The efPCs that are important for the test data may be different than those that are important for the training data. If there is a difference, it may suggest that test data contains observations that are out of distribution compared to the training data.
\item \emph{Visualization of efPCs} Create visualizations of the principal directions for any important efPCs not previously considered.
\end{enumerate}

We apply the above steps to the shifted peaks test data functions using
the random forest model described in Section \ref{pipeline-train}. The
test data accuracy is 0.99. The resulting PFI values on the test data
are depicted in Figure \ref{fig:fig3} (bottom). Note that the magnitude
of the importance is larger for the test data than the training data
(i.e., a PFI value of 0.033 for jfPC 1 on the test data compared to a
PFI value of 0.001 for jfPC 1 on the training data). It would be
interesting to explore this further to better understand the reasons for
the magnitude difference. In the training data, only jfPC 1 has
importance, but with the test data, both jfPC 1 and 2 show importance.
There are also several PCs with negative PFI values on the test data,
which suggests that the model performs better with these variables
removed.

\begin{figure}

{\centering \includegraphics[width=4in]{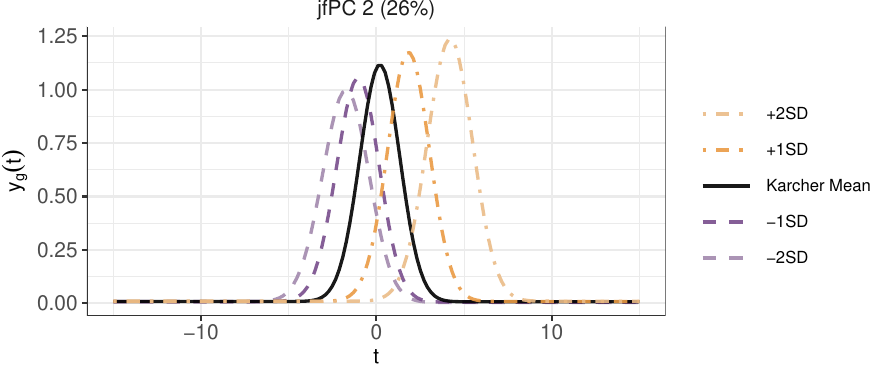} 

}

\caption{Plot of the principal directions for interpreting the functional variability captured by jfPC 2 from the shifted peaks data shown in Figure \ref{fig:fig1}.}\label{fig:fig4}
\end{figure}

Recall that Figure \ref{fig:fig2} shows the principal directions
associated with jfPC 1, which indicated that jfPC 1 captures joint
variability in peak time and peak intensity. Figure \ref{fig:fig4} shows
the principal directions associated with jfPC 2. jfPC 2 appears to
capture both horizontal and vertical variable between functions that is
more nuanced that jfPC 1: functions that are similar to the Karcher mean
and functions that are different. Both of these aspects of the
variability in the shifted peaks functions makes sense to be useful for
classifying the two groups.

\subsection{\texorpdfstring{Intended Use
\label{purpose}}{Intended Use }}\label{intended-use}

The development of explanations for machine learning models is dictated
by the intended audience. For example, an explanation intended for a
model developer who is knowledgeable about the technical details of a
model may be more technical. However, an explanation intended to promote
the use of a model to a decision maker (e.g., a medical doctor or
courtroom jury member) will likely need to take a less technical form
for an individual who is not familiar with the technical details of the
model. In this section, we clarify that the intended use for the
explanations from the VEESA pipeline are intended to be used by analysts
for model development. Additional work is needed to determine useful
approaches for distilling the information gained from the VEESA pipeline
to decision makers who are not knowledgeable about the technical details
of machine learning models. However, anecdotal experience has shown us
that the resulting explanations from the VEESA pipeline can also be
useful to the application subject matter experts since the explanations
are presented in the data space that the experts are used to working
with. This allows the subject matter experts to connect the functional
variability identified as important to their scientific understanding of
the underlying mechanism.

The intended uses for the explanations produced by the VEESA pipeline
are:

\begin{itemize}
\setlength{\itemindent}{1em}
\item \emph{Variable Selection}: At this point in the model development process, an analyst can use the PFI results to help with variable selection. PFI determines the efPCs that are globally important to a model for prediction. Thus, a new model can be trained using a reduced number of efPCs selected based on PFI. The efPCs with low variable importance will likely capture variability created by noise in the data or factors unrelated to the response variable. The removal of these efPCs from a model may improve predictive performance.
\item \emph{Model Comparison}: If multiple models are trained (including different model types), the VEESA pipeline helps analysts determine if the models are using similar or different characteristics of the functional variability. If the characteristics are different, the information may help in selection of a model. For example, a model that uses the most reasonable variables based on the scientific understanding of the application may be more desirable since the predictions can be explained. Another scenario is that a model that has fewer important principal components with comparable predictive performance may be more desirable for easier interpretability.
\item \emph{Gaining Model Trust}: Model trust is built through many aspects (e.g., predictive performance, code accuracy, etc.), and one such aspect is whether the data are being utilized by the model in a scientifically reasonable manner. The VEESA pipeline allows analysts to identify the aspects of functions that are used by a model for discrimination and determine if the functional variability important to the model is reasonable based on an understanding of the phenomenology of the data. Trust is gained in the model if reasonable data characteristics are used for discrimination. For example, the PFI results from the shifted peaks data indicate that the random forest is using peak height and timing to classify function to a group. Since we know that this is the main phenomenological characteristic that distinguishes the two classes, we gain trust in the model. If the PFI results indicated that only jfPCs that captured vertical variability were important, we would have been suspicious of the model not accurately capturing the appropriate information in the data. Of course, we know the true generating mechanism for the shifted peaks data, which will not be the case with real data. However, subject matter expert knowledge could be used to inform whether the model is considering reasonable data characteristics and/or missing data characteristics that should be informative.
\end{itemize}

\section{\texorpdfstring{Examples
\label{examples}}{Examples }}\label{examples}

In this section, we present two examples that utilize the VEESA pipeline
to train machine learning models with functional data inputs and gain
insight into the models. Section \ref{hct} applies a neural network to
identify an explosive from a set of materials using hyperspectral
computed tomography (H-CT) scans. Section \ref{ink} applies a random
forest to predict inkjet printer source using Raman spectroscopy. Both
examples are high consequence applications, where it is important to
provide reasoning for the predictions for the models to be trusted to
make decisions. The models selected for these analyses are selected as
examples of two different commonly used black-box machine learning
models. The data in these examples are available upon request, and the
code is available at
\url{github.com/sandialabs/veesa/tree/master/demos/goode-et-al-paper}.

\subsection{\texorpdfstring{Hyperspectral Computed Tomography Data
Material Classification
\label{hct}}{Hyperspectral Computed Tomography Data Material Classification }}\label{hyperspectral-computed-tomography-data-material-classification}

\begin{figure}

{\centering \includegraphics[width=5.5in]{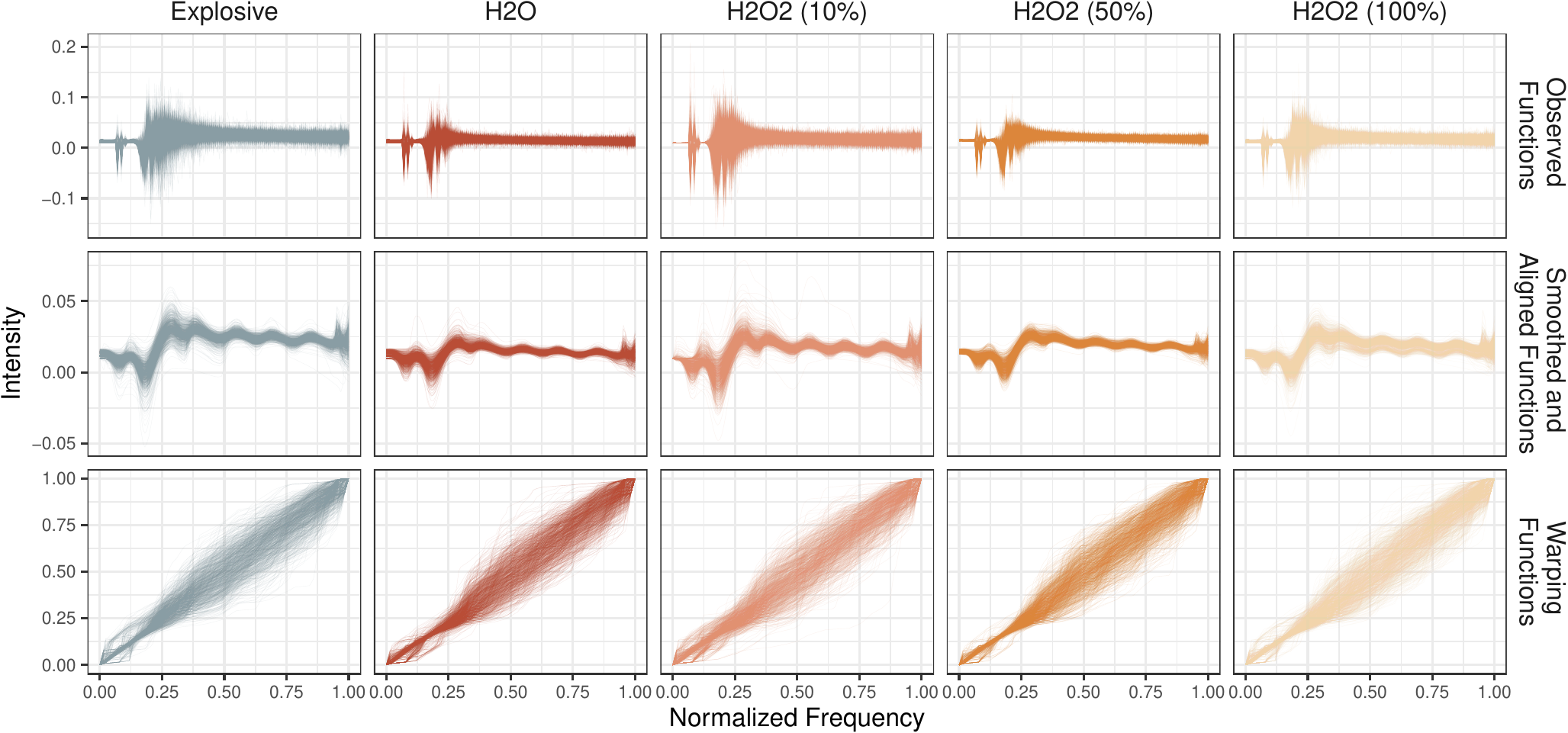} 

}

\caption{Observed (top row), smoothed and aligned (middle row), and warping functions (bottom row) of a subset of 1,000 H-CT signatures for each material.}\label{fig:fig5}
\end{figure}

Hyperspectral computed tomography (H-CT) scans of materials produce a
signature across a set of frequencies unique to an observation. There is
interest in using H-CT scans to identify explosives from a set of
materials for applications such as airport security
\citep{jimenez:2017, gallegos:2018}. In this example, we consider a set
of 1,980,409 H-CT scans simulated by experts at Sandia National
Laboratories that contain five materials: water (\(H_2O\)), hydrogen
peroxide (\(H_2O_2\)) solutions diluted by water with 100\%, 50\%, and
10\% \(H_2O_2\), and an explosive \citep{gallegos:2019}. The percentages
of observations per material are 34\%, 17\%, 8\%, 7\%, and 34\%,
respectively. The signatures are separated into training and testing
sets by randomly sampling 80\% of scans from each material to be
included in the training data. The top row of Figure \ref{fig:fig5}
shows subsets of 1,000 randomly selected signatures per material from
the training data. The scans record observations at 128 frequencies,
which are visualized as normalized frequencies between 0 and 1.

We apply the VEESA pipeline with a neural network as the model to
predict the material of an H-CT scan. Smoothing is performed using a
box-filter. The number of times the box filter is run (15 times) is
determined by considering a range of values and selecting the value with
the best predictive performance on the test data. The details of the
smoothing process are included in the supplemental material. Figure
\ref{fig:fig5} (middle and bottom rows) shows the subset of 1,000
signatures after smoothing and alignment and the corresponding warping
functions.

\begin{figure}

{\centering \includegraphics[width=5in]{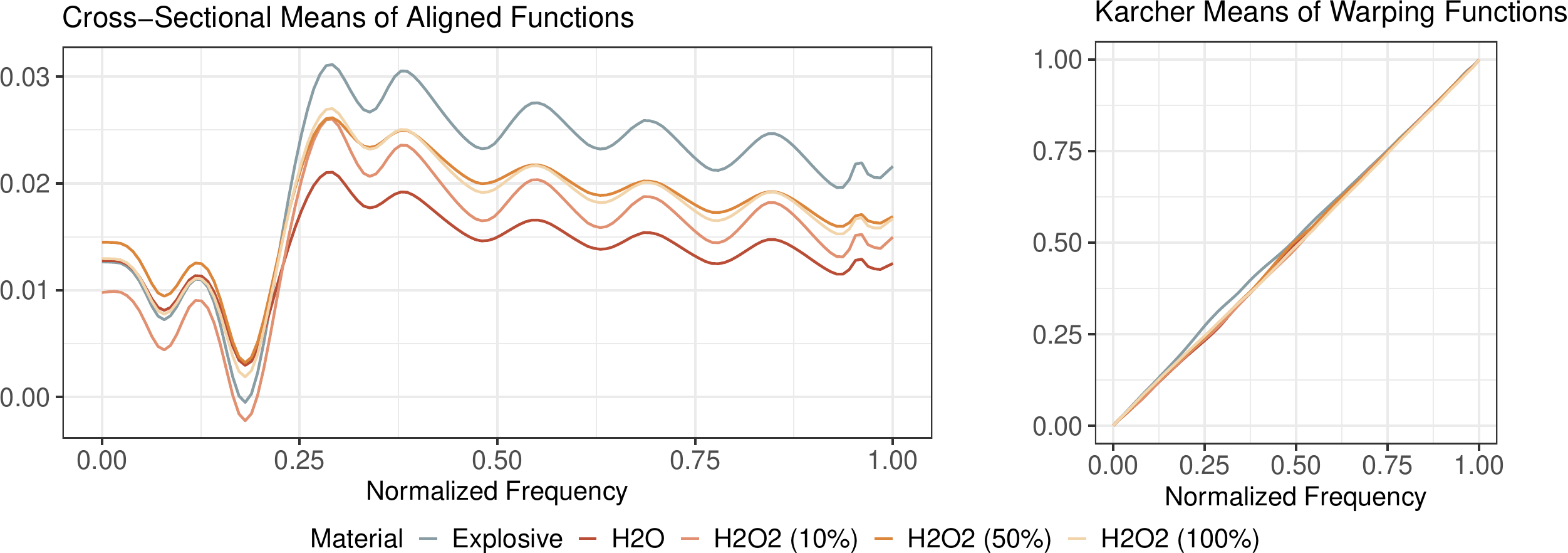} 

}

\caption{(Left) Cross-sectional functional means of the aligned functions in Figure \ref{fig:fig5} for each material. (Right) Karcher means of the warping functions within each material.}\label{fig:fig6}
\end{figure}

As an exploratory analysis step, we compute the cross-sectional means of
the aligned signatures for each material and the Karcher means of the
warping functions for each material (Figure \ref{fig:fig6}). The aligned
means show clear vertical variability between the materials, and the
pattern in vertical variability changes just before the normalized
frequency of 0.25. For example, water and the explosive have similar
intensities before a normalized frequency of 0.25. After the change, the
explosive has the highest intensity, and water has the lowest intensity.
There are minimal differences between the warping function material
means, which indicates there is little horizontal variability between
the H-CT functions, on average.

The trends in Figure \ref{fig:fig6} suggest that the characteristics in
the data for discriminating between materials is likely captured in the
vertical variability. As a result, we hypothesize that vfPCA will be a
better option for discriminating between materials, but we also consider
jfPCA in case there is discriminatory information in the horizontal
variability that is not captured by the means. We train two models: one
with vfPCs as inputs and one with jfPCs as inputs. For both models, all
128 PCs are included. The models are trained using the Python package
\emph{scikit-learn} \citep{pedregosa:2011} with all default values
(i.e., one layer with 100 neurons and a ReLU activation function). We
elect to not perform model tuning for this analysis since PFI provides
insight into a model regardless of whether it has good or bad predictive
performance. Instead, we fit these two models and work to understand how
they make use of the data. Additional work could be done to use
information gained from this initial application of the VEESA pipeline
to improve a model. PFI is applied to the test data using 5 replications
and a performance metric of accuracy. A comparison of applications of
the VEESA pipeline with a cross-sectional approach is included in the
supplemental material.

\begin{figure}

{\centering \includegraphics[width=5in]{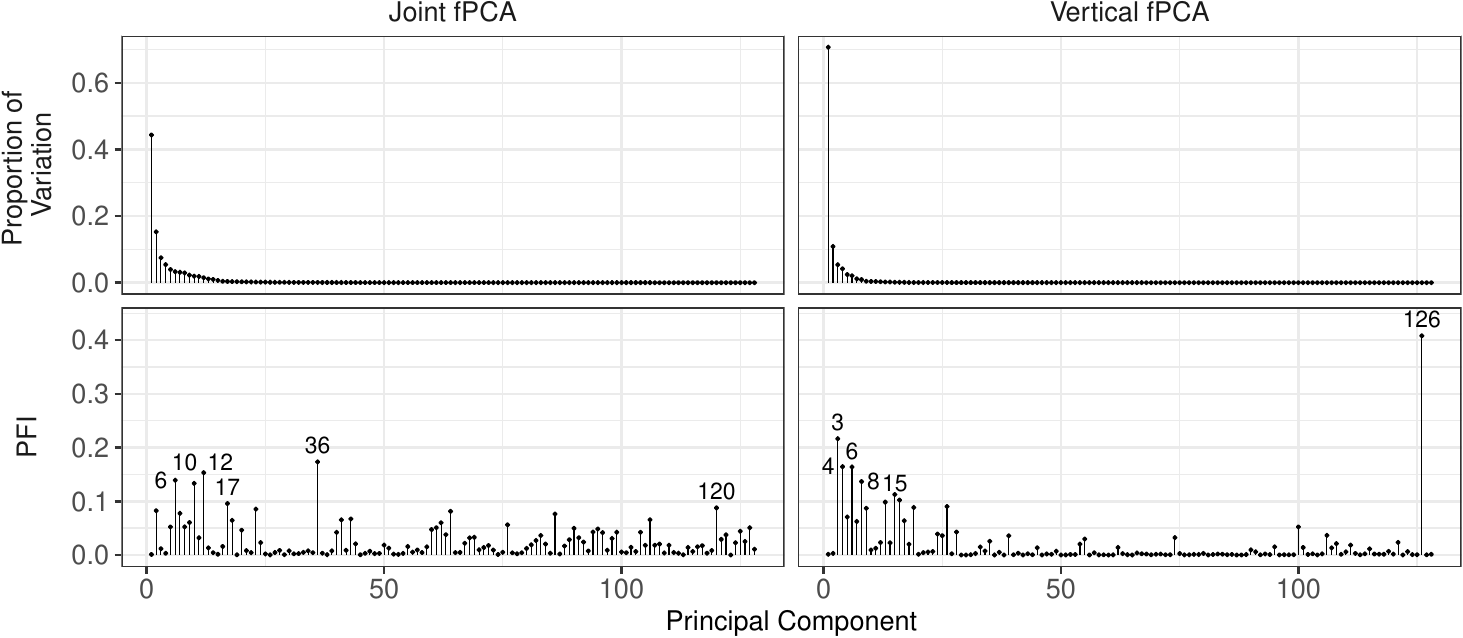} 

}

\caption{Proportion of variation (top) and PFI values (bottom) associated with the jfPCs (left) and vfPCs (right) in the H-CT example. The four jfPCs and vfPCs with the highest PFI are labeled.}\label{fig:fig7}
\end{figure}

Figure \ref{fig:fig7} shows the proportion of variation and PFI value
associated with the jfPCs (left) and vfPCs (right). The variability
across PFI replicates is not depicted since it is much smaller than the
variability across PCs; see the supplemental material for more details
on the variability. For both the vfPCA and jfPCA models, the proportion
of variation has a major drop after the first PC and drops close to zero
after a small number of PCs. In contrast, there is a much different
relationship between PFI values and principal component number, which
provides a clear example where predictive value of a principal component
is not directly related to the amount of variability explained by a
principal component. For the vfPCs, PC 126 has an extremely high PFI
value and a handful of earlier PCs have elevated PFI values. With the
jfPCs, 4 PCs (before PC 50) stand out with PFI values higher than 0.1.

As expected, the model fit with vfPCs returns a higher test data
accuracy of 0.88 compared to the model fit with jfPCs with a test data
accuracy of 0.81. We consider some of the principal directions from the
better performing vfPC model here and include principal directions from
the jfPC model in the supplemental material. Figure \ref{fig:fig8}
depicts the principal direction plots of the 6 vfPCs with the highest
PFI values (i.e., the PCs with PFI values above 0.1). The PC with the
highest PFI (vfPC 126) captures a consistent vertical variability across
all frequencies. The other five vfPCs capture more nuanced aspects of
the variability. vfPC 3 captures the contrast in intensities before and
after frequency 0.25. vfPC 4 focuses on scans with low intensities
around frequencies of 0.2 and 0.35 and above average intensities after
0.35. vfPCs 6, 8, and 15 all capture the change in intensities before
and after 0.2 with slight different focuses: vfPC 6 focuses on the first
peak after 0.2, vfPC 8 considers functions close to the mean function
between approximately 0.2 and 0.3, and vfPC15 considers functions close
to the mean function between approximately 0.7 and 0.85.

The most important vfPCs appear to capture characteristics in the H-CT
data that are reasonable given the clear distinction between the aligned
material means in Figure \ref{fig:fig6}. This is a promising start for
an analysis of the H-CT. Next steps could involve removing vfPCs with
PFI values close to 0 and model hyperparameter tuning to see if test
data accuracy improves.

\begin{figure}

{\centering \includegraphics[width=5.5in]{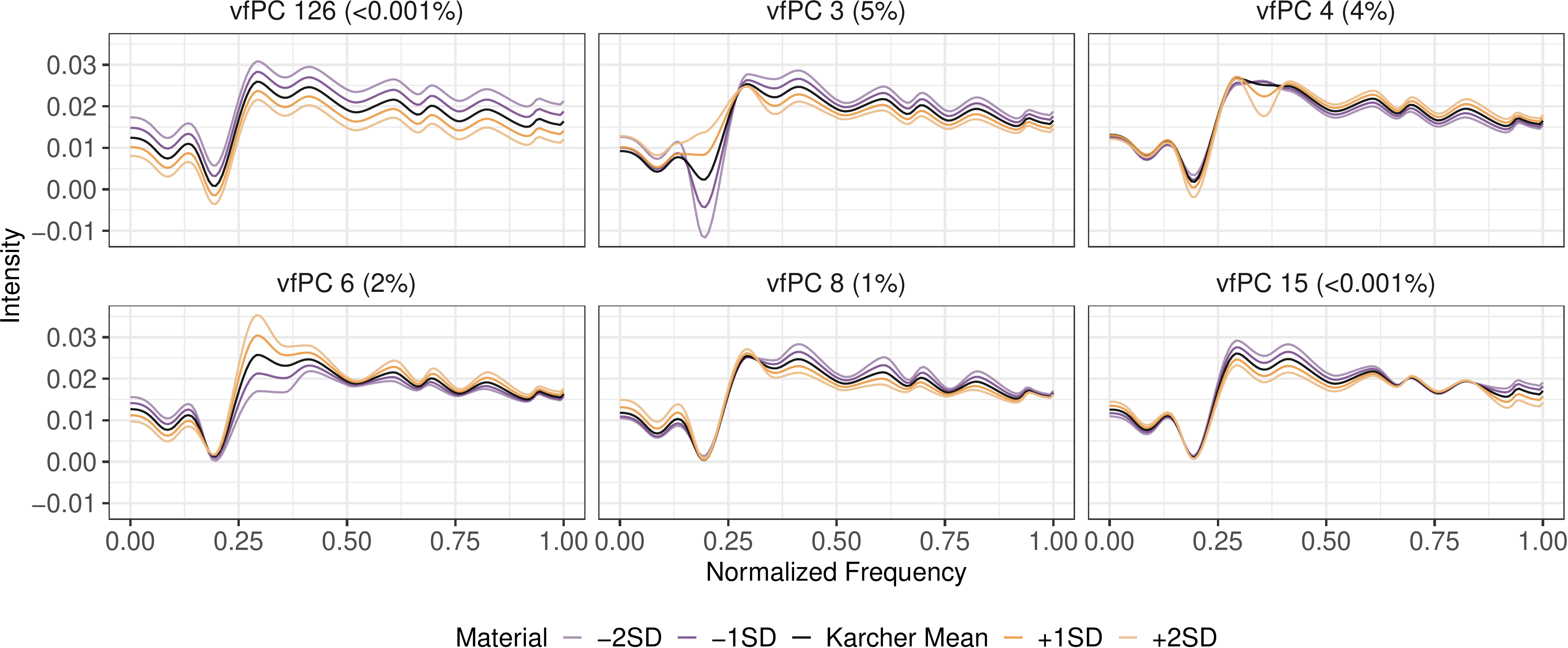} 

}

\caption{Principal directions from the six vfPCs with the highest PFI from the H-CT data example.}\label{fig:fig8}
\end{figure}

\subsection{\texorpdfstring{Inkjet Printer Identification with Raman
Spectroscopy
\label{ink}}{Inkjet Printer Identification with Raman Spectroscopy }}\label{inkjet-printer-identification-with-raman-spectroscopy}

\begin{figure}

{\centering \includegraphics[width=5.5in]{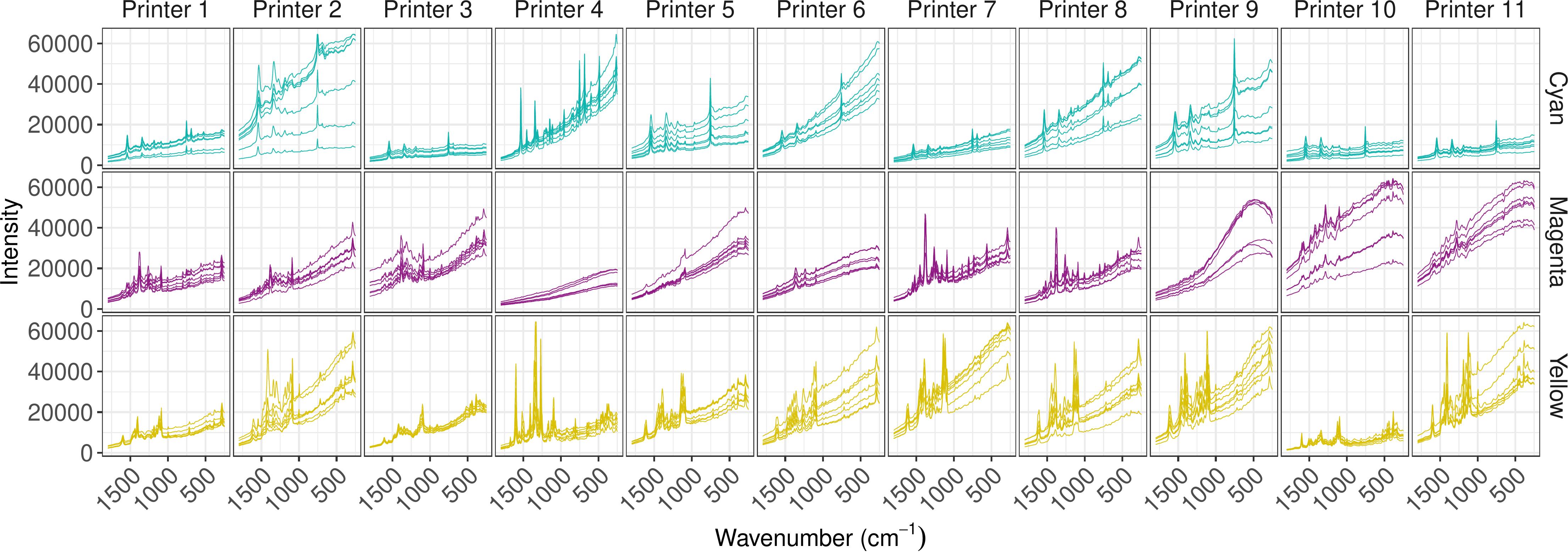} 

}

\caption{Raman spectra signatures from 11 inkjet printers for the colors of cyan, magenta, and yellow.}\label{fig:fig9}
\end{figure}

Inkjet printers can be used for illicit activities such as printing
counterfeit currency. As a result, forensic investigators are interested
in techniques that provide evidence connecting printed material to the
source printer. \citet{buzzini:2021} present one approach where Raman
spectroscopy is used to extract signatures from documents generated by
inkjet printers. The authors then use different variants of linear
discriminant analysis (LDA) to predict the source printer given a sample
of Raman spectra, which we will refer to as a signature. Figure
\ref{fig:fig9} shows signatures from different printers that
\citet{buzzini:2021} considered for their analyses. We will refer to
this set of samples as the \emph{inkjet dataset}. Since the Raman
spectra signatures are functional data, we are interested in applying
the VEESA pipeline to predict the source printer for a document given a
Raman spectra signature and understand the functional variability
important to the model for prediction.

The inkjet dataset (Figure \ref{fig:fig9}) contains signatures collected
from eleven documents belonging to the Counterfeit Forensic Section of
the Criminal Investigative Division of the US Secret Service. Each
document is printed from a different device, but some of the devices
have the same manufacturer as shown in Table 2. In addition to sharing a
manufacturer, printers 7 and 8 also are the same model of printer. For
each document, 7 replicates were collected from the three main colored
dot components (cyan, magenta, and yellow). Each replicate signature
contains observations at 231 spectra between 1800 and 250 cm\(^{-1}\).
\citet{buzzini:2021} converted the replicates to have 1129 observations
per signature with a path of approximately 1.4 cm\(^{-1}\). See
\citet{buzzini:2021} for more details on the collection process.

We choose to implement the VEESA pipeline using jfPCA since Figure
\ref{fig:fig9} shows that there is variability in both the vertical and
horizontal directions in the signatures, and we do not have prior
knowledge of which type of variability contains information useful for
prediction. \citet{buzzini:2021} compare various preprocessing methods
applied to the signatures before applying LDA. Their preprocessing steps
included a baseline correction and different forms of normalization. We
elect to only implement a box-filter for the smoothing step in the VEESA
pipeline for preprocessing the functions before applying jfPCA. The
variability removed by a baseline correction and normalization should be
captured by the the jfPCs, and then the model should identify which
modes of variability captured by the jfPCs are useful for prediction. We
are interested in determining if we can achieve equal or better
performance without preprocessing.

We select a random forest as the predictive model as an example of
commonly used statistical learning model. We follow a similar set up as
\citet{buzzini:2021} for the model predictive assessment, so we can
compare our model performance to their best performing models.
Specifically, we build predictive models separately for each color, and
we implement a 3-fold cross validation procedure for predictive
performance assessment. The folds are created such that each replicate
for a printer and color is randomly assigned to be in fold 1, 2, or 3
(with 3 functions in fold 1 and 2 functions in folds 2 and 3). Then two
of the folds are used to train a model using Steps 1-4 of the VEESA
pipeline described in Section \ref{pipeline-train}, and the third fold
is used for testing Steps 1-4 of the VEESA pipeline described in Section
\ref{pipeline-test}. The procedure is repeated such that the three
possible fold pairs are used for training. Since the fold a signature is
assigned to is random, this 3-fold cross validation procedure is
repeated a total of 10 times to account for variability.

\begin{longtable}[]{@{}cll@{}}
\caption{Printer manufactures and models in the inkjet
dataset.}\tabularnewline
\toprule\noalign{}
Printer & Manufacturer & Model \\
\midrule\noalign{}
\endfirsthead
\toprule\noalign{}
Printer & Manufacturer & Model \\
\midrule\noalign{}
\endhead
\bottomrule\noalign{}
\endlastfoot
1 & Brother & MFC-665CW \\
2 & Canon Pixma & MX340 \\
3 & Canon PG & 210XL \\
4 & Epson & Unknown \\
5 & HP & Officejet 5740 \\
6 & HP & Deskjet f5180 \\
7 & HP & Officejet 6500 \\
8 & HP & Officejet 6500 \\
9 & Lexmark & 228 2010 CE 81 \\
10 & Sensient & Unknown \\
11 & Sensient & Unknown \\
\end{longtable}

For each color, model accuracy is computed for each test fold and
repetition, and the cross validation predictive performance metric is
computed as the average of the test-fold accuracies across the 10
replicates. That is, let \(y_i\) represent the true printer associated
with observation \(i=1,...,77\), and let \(\hat{y}_{i,r}\) represent the
predicted value of \(y_i\) when observation \(i\) is in the test fold
for replicate \(r=1,...,10\). The cross validation average accuracy for
one color (\(color\in\{\mbox{cyan},\mbox{magenta},\mbox{yellow}\}\)) is
computed as
\[Acc_{color} = \frac{1}{10\cdot77}\sum_{r=1}^{10}\sum_{i=1}^{77}I[y_i=\hat{y}_{(i,r)}].\]

There are a handful of values that must be specified when implementing
the VEESA pipeline. We choose to vary the number of times the box-filter
is run for smoothing (0, 5, 10, 15, 20, 25, 30, and 35 times), the
number of PCs input to the model (10, 20, 30, 40, 50, 60, 70, 80, 90,
and 100 PCs), and the number of trees in the random forest (50, 100,
250, 500, and 1000 trees). The cross validation process is applied 400
times for each color (once for all combinations of box-filter runs,
input PCs, and random forest trees). The values considered are selected
to show a range over which model accuracy increases and diminishes. The
random forests are fit using the \emph{randomForest} R package
\citep{liaw:2022} with all tuning parameters set to the default value
besides for the number of trees.

\begin{figure}

{\centering \includegraphics[width=5.5in]{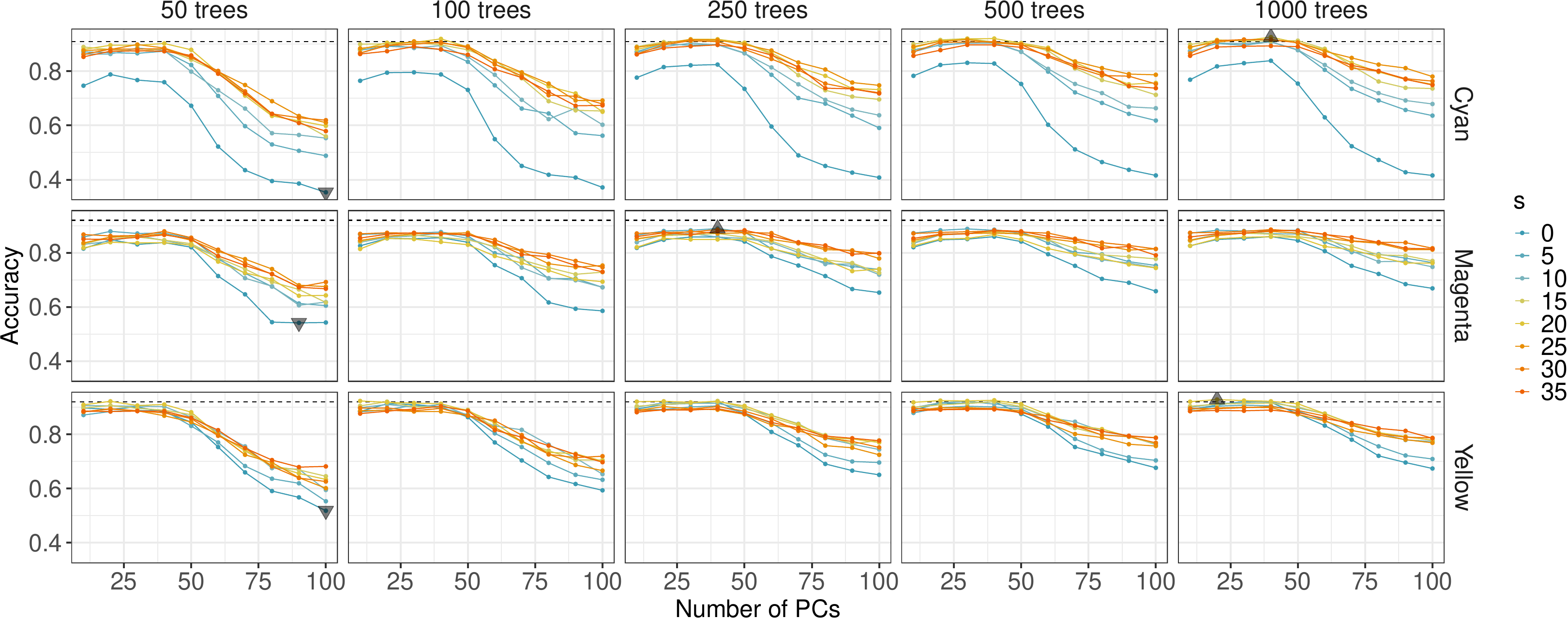} 

}

\caption{Cross validation average accuracies. Triangles pointing up and down highlight the highest and lowest average cross validation accuracies from the VEESA pipeline, respectively, for each color. Horizontal dashed lines represent best average cross validation accuracies for each color from \citet{buzzini:2021}.}\label{fig:fig10}
\end{figure}

The cross validation average accuracies are shown in Figure
\ref{fig:fig10}. The x-axis shows the number of PCs input to the model,
and the y-axis depicts the cross validation accuracy. The rows and
columns separate the results by color and the number of random forest
trees, respectively. The color of a line represents the number of times
the box filter is run, and the horizontal dashed black line represents
the best cross validation accuracy for a color obtained by
\citet{buzzini:2021}. There are some clear trends in the cross
validation results. For example, regardless of the number of random
forest trees, the average accuracies tend to increase as the number of
PCs increases to around 30-50 and then decrease as the number of PCs
increase further. We also see that for all PCs, number of trees, and
color, the more times the box-filter is run, the average accuracy tends
to increase until approximately 25 times when the improvement is either
minimal or the accuracy begins to decrease.

The grey triangles pointing up and down in Figure \ref{fig:fig10}
indicate the VEESA pipeline scenarios with the highest and lowest
average cross validation accuracy, respectively, within a color. Table 3
contains the number values of the highest and lowest CV averaged
accuracies associated with the triangles in Figure \ref{fig:fig10}. The
values associated with the VEESA pipeline implementation that results in
these values are also included (i.e., number of PCs, number of trees,
and smoothing iterations). Note that the lowest accuracies from the
VEESA pipeline have no smoothing and many PCs. For cyan and yellow, we
achieve approximately the same or better predictive performance. For
magenta, our highest average cross validation accuracy is close but
lower. Additional investigations show that the low magenta accuracy with
the VEESA pipeline is due to printers 7 and 8 being classified as the
other printer. These two printers have the same manufacturer and model,
and Figure \ref{fig:fig9} shows that their signatures similar.
Additional details about individual printer classifications are included
in the supplemental material.

\begin{longtable}[]{@{}llrrrrl@{}}
\caption{Cross validation average accuracies from the best and worst
performing VEESA pipeline models applied to the inkjet dataset. The last
column contains the highest cross validation accuracies achieved by
\citet{buzzini:2021}.}\tabularnewline
\toprule\noalign{}
Scenario & Color & Box Filter & PCs & Trees & VEESA & Buzzini \\
\midrule\noalign{}
\endfirsthead
\toprule\noalign{}
Scenario & Color & Box Filter & PCs & Trees & VEESA & Buzzini \\
\midrule\noalign{}
\endhead
\bottomrule\noalign{}
\endlastfoot
Best & Cyan & 20 & 40 & 1000 & 0.9260 & 0.91 \\
Best & Magenta & 5 & 40 & 250 & 0.8896 & 0.92 \\
Best & Yellow & 20 & 20 & 1000 & 0.9286 & 0.92 \\
Worst & Cyan & 0 & 100 & 50 & 0.3532 & - \\
Worst & Magenta & 0 & 90 & 50 & 0.5416 & - \\
Worst & Yellow & 0 & 100 & 50 & 0.5182 & - \\
\end{longtable}

For each of the best and worst performing tuning parameter scenarios, we
train a random forest on the data from all printers. Again, the models
are trained separately for each color. We elect to train these models on
all data within a color to mimic the joining of all available data to
build a model for predictions on new data without a known printer. We
apply the VEESA pipeline to understand how such a model makes use of the
functional variability in the inkjet signatures. We choose to consider
both the best and worst cases to study how the feature importance varies
between good and poor performing models.

\begin{figure}

{\centering \includegraphics[width=5.5in]{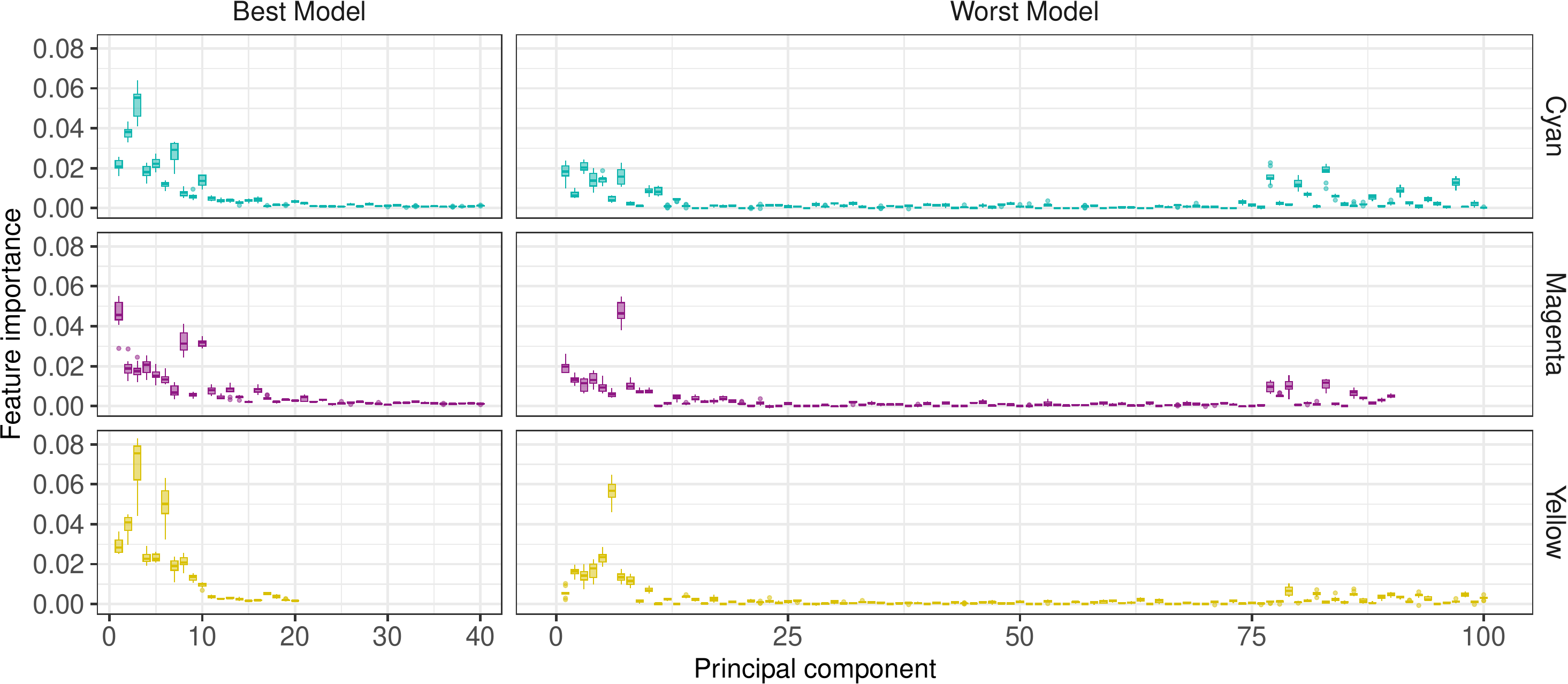} 

}

\caption{Boxplots of PFI values across 10 replications from the best and worst performing models for cyan, magenta, and yellow inkjet signatures.}\label{fig:fig11}
\end{figure}

When implementing the VEESA pipeline with all signatures for a color, we
apply all steps described in Section \ref{pipeline-train} with jfPCA. We
compute PFI for the jfPCs using 10 replications. The PFI results are
included in Figure \ref{fig:fig11}. The best performing models place
high importance on the earlier PCs, and the worst performing models
place importance on both early PCs and later PCs (i.e., the PCs with
high are either less than 15 or greater than 75). The low importance of
PCs 15 to 75 suggest that they may be able to be removed to improve
model predictive performance of the worst performing scenarios. We
explore how the removal of these PCs affects the model accuracies in the
supplemental material.

\begin{figure}

{\centering \includegraphics[width=5.5in]{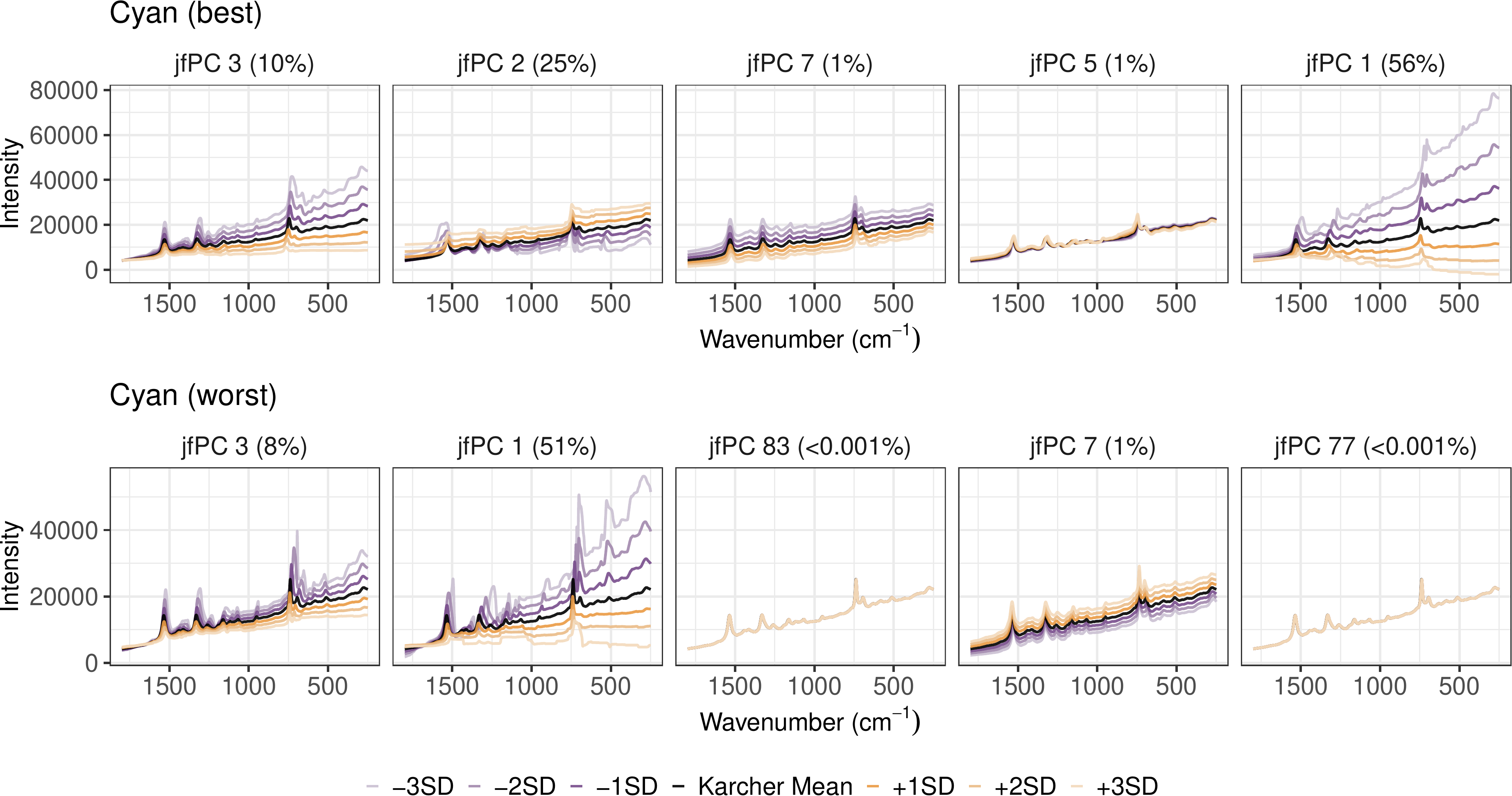} 

}

\caption{Principal directions from the best (top row) and worst (bottom row) models for predictions with cyan inkjet signatures. The jfPCs selected are those with the largest PFI values for their respective model. PCs are ordered from left to right based on highest to lowest feature importance.}\label{fig:fig12}
\end{figure}

Figure \ref{fig:fig12} shows principal directions from the best and
worst performing models for cyan signatures. For space reasons, we only
consider the five PCs with the highest feature importance values. We
start by considering the important principal directions associated with
the best model (Figure \ref{fig:fig12} top). While there are nuances in
the variability captured, jfPCs 1, 2, and 3 generally capture vertical
variability with increasing variability as the wavenumber approaches
500. Each of these PCs also captures small amounts of horizontal
variability. For example, the most important PC, jfPC 3, captures small
amounts of horizontal variability in the first two main peaks that occur
at larger wavenumbers than the mean. For the third peak, jfPC 3 captures
variability in functions that peak before and after the mean. jfPC 7
consistently captures approximately the same amount of vertical
variability across wavenumbers. Lastly, jfPC 5 captures small amounts of
vertical variability before and after the third main peak. It seems
reasonable that the model would focus on these modes of variability for
prediction. In Figure \ref{fig:fig9}, a key distinguishing factor
between printers in cyan signatures is the different in the slope of the
intensity values as the wavenumbers approach 500 cm\(^{-1}\). Another
distinguishing factor is the height of the main three peaks and small
amounts of variability in the wavenumbers where the main peaks occur.

The most important jfPCs for the worst performing model capture similar
aspects of functional variability as the best performing model. In fact,
the jfPCs of 1, 3, and 7 appear again. Even though no smoothing is
applied to the signatures with this model, the larger picture functional
variability matches that captured by the jfPCs with the same numbers.
However, these PCs are clearly less smooth. The other two important
jfPCs that fall in the top five (jfPCs 77 and 83) capture such little
variability that it is not visible in these figures. Instead, these
capture variability in functions that are similar to the mean, versus
those that are not. It is possible that information could be useful for
classification (i.e., some of the printer signatures in Figure
\ref{fig:fig9} are similar to the Karcher mean and some are very
different). However, the additional noise in the functions when no
smoothing is applied must make it difficult to separate key
characteristics of the signatures that are unique to a printer.

The principal directions associated with the most important PCs for the
best and worst magenta and yellow models are included in the
supplemental material. Overall, we are able to achieve similar
predictive performance as the analysis in \citet{buzzini:2021} while
additionally gaining insight into how the random forests make use of
patterns in the data. We are able to achieve this predictive performance
without as much preprocessing of the data as implemented in the analysis
of \citet{buzzini:2021}. We point this out, because this also highlights
that the VEESA pipeline could be beneficial for analyses that do not
have access to subject matter experts or scenarios where it is not
previously understood how to best preprocess the data based on prior
experiences.

\section{\texorpdfstring{Discussion
\label{discussion}}{Discussion }}\label{discussion}

While there has been much research on explainable machine learning,
there is little research that specifically considers functional data as
model inputs and accounts for the dependence structure in the data. The
VEESA pipeline provides a way for model developers to train a supervised
machine learning model with functional data inputs that (1) accounts for
functional dependence and the horizontal and vertical functional
variability of functional data inputs and (2) identifies the aspects of
functional variability that are globally important to the model. This is
done by first applying some pre-processing steps: the functions are
smoothed, the horizontal and vertical variability in the functions are
separated using the ESA framework, and efPCA is applied. Then a machine
learning model is trained with the efPCs as the predictor variables, and
PFI is used to identify the efPCs important for prediction. Finally, the
principal directions of the important efPCs are visualized to interpret
the functional variability captured. The process may also be applied to
test data to gain insight into how the model produces predictions on
data not seen during training. The VEESA pipeline offers one approach to
explainable machine learning with functional data, but this approach is
only a start. There are numerous directions for improvement and future
research.

An important element to consider, as previously discussed, is the nature
of explanations is dictated by the intended audience. The VEESA pipeline
is aimed at providing insight to model developers and, potentially,
subject matter experts. However, if a model developed using the VEESA
pipeline is deployed, additional steps should be taken to determine the
form of explanations aimed at users or decision makers who do not
possess the same technological knowledge as the model developers or
subject matter experts. The manner in which a model would be used in
practice could take different forms. For example, in the inkjet
analysis, a model developer may apply the model to a new prediction that
could be used in a court case, but the developer may need to be creative
to find a way to distill their understanding of what is important to the
model to provide explanations to lawyers, judges, and juries. In the
H-CT material classification, consider the scenario where a model
developed with the VEESA pipeline is used by airport security to scan
luggage for explosives. Simply providing the global PFI values will not
be sufficient to help the security employees trust and work with the
model. Instead, an interactive tool that allows the employees to explore
understandable explanations for individual predictions (i.e., local
instead of global explanations) would be more valuable (e.g.,
\citet{wang:2022}). In general, there is much room for the advancement
of explanations intended for decision makers as highlighted by
\citet{bhatt:2020}.

As briefly mentioned in the previous paragraph, there are two types of
explanations: global and local. Global explanations aim to provide a
summary of how a model makes use of data across an entire set of
predictions. Local explanations aim to provide reasoning for how a model
makes use of data for an individual prediction. PFI is a global
explanation technique. Other types of global explanations include
partial dependence plots (PDPs) \citep{friedman:2001} and global
surrogate models \citep{molnar:2022}. Local methods include individual
conditional expectation (ICE) plots \citep{goldstein:2015} and SHAP
\citep{lundberg:2017}. Any of these or other explainability techniques
could be used in place of PFI in the VEESA pipeline. For example, PDPs
would provide a visual of the average marginal relationships between the
predictions and the efPCs. ICE plots would provide the same visual but
for individual predictions. Arguably, the best route would be to draw on
the explainability tool box and apply multiple methods to gain different
perspectives into the model. In addition to efPCs accounting for both
types of variability present in functional data, the use of efPCs
provides an additional advantage to using other explainability methods.
Just as PFI is known to provide biased results when correlation is
present between variables, PDPs, ICE plots, SHAP, and other
explainability techniques are also known to be negatively affected by
correlation between predictor variables
\citep{molnar:2020b, hooker:2021}. Since the efPCs are uncorrelated,
explainability methods known to be negatively affected by correlation
are able to be applied without concern due to correlation.

There are also various directions for research specific to PFI. We
describe these topics in the context of the VEESA pipeline, but the
ideas apply more generally. If PFI returns a large number of efPCs with
significant importance, there may be too much information for analysts
to extract a clear understanding. Future work could investigate an index
for specifying whether a model has high or low explainability
potentially based on the number of important features and their
usefulness. Such a metric could help determine what level of deployment
a model is ready for depending on the consequential nature of the
application. Additionally, future work could investigate a method for
determining a ``cut off'' for which efPCs are considered important,
which could be useful for feature selection (i.e., remove the
unimportant or less important efPCs, retrain the model, and determine if
interpretability is improved with fewer efPCs while maintaining or
improving on predictive performance).

Another issue is the computational intensity of PFI. For large machine
learning data sets, the process of applying PFI to a dataset may need to
be adjusted. One route may be to select a subset that is representative
of the data of interest that PFI is applied to. A survey sampling method
may be applicable here. Instead of a representative sample, a subset
could also be chosen based on observations of interest, which may
provide valuable insight to a model regardless of the data size. For
example, PFI could be applied separately to clusters identified in the
data, only observations that are predicted incorrectly by the model, or
separately by classes. Considering feature importance separately by
class may be of special interest in the case of uneven class sizes. A
different route to take with large datasets to apply PFI to all
observations would be to break the computation into pieces that could be
implemented independently and later combined. Other areas of active
research with feature importance include quantifying variable
interactions (e.g., \citet{greenwell:2018}) and computing uncertainty
(e.g., \citet{williamson:2023}).

There are several unspoken assumptions underlying the VEESA pipeline
that should be addressed. First, we assume that the efPCs have
meaningful contextual interpretations. If the efPCs are difficult to
interpret, then it will not be clear what aspects of the functional
variability are driving the model predictions. A way to resolve this
issue is to use a different method for summarizing the functions while
appropriately accounting for their variability such as a varimax
rotation \citep{ramsay:2005}. It may also be possible to substitute
efPCA with a different method that accounts for both the horizontal and
vertical variability, is interpretable, and provides uncorrelated
variables. Some possible methods include partial least squares (PLS) for
functional data \citep{febrero:2017} and canonical correlation analysis
\citep{lee:2017a}.

The second assumption we make is that the only inputs to the model are
efPCs from one set of functions. If there are additional variables
(e.g., individual vectors or efPCs from another set of functions), the
guarantee of uncorrelated input variables may no longer hold. For
example, in the inkjet analysis, a clear next step is to train a model
including all three colors. However, if the efPCs computed separately on
cyan, magenta, and yellow signatures are included, the PFI may be
biased. There is some work on computing feature importance that accounts
for correlation between input variables (e.g., \citet{strobl:2008} and
\citet{hooker:2021}, but this is another area of active research.

The final assumption is that the approach for implementing efPCA
described in this paper assumes that the functions have the same number
of peaks. If the functions have different numbers of peaks, the approach
has to decide which nearby peak to align a function with, which could
affect classification. Future work could explore the use of a Bayesian
multimodal alignment approach that has been proposed for handling this
scenario under the ESA framework \citep{tucker:2021}.

The VEESA pipeline is an example of a procedure that combines methods
commonly thought of as ``statistical'' methods (efPCA) with machine
learning approaches (PFI). By joining statistical and machine learning
approaches, the pipeline takes advantage of the benefits of both: the
ability to capture the data dependence structure with efPCA and the data
driven predictive ability of machine learning. We echo previous authors
that are encouraging more meshing of the two cultures in hopes of the
development of new approaches that draw on the benefits of the two
approaches.

\section*{Acknowledgements}\label{acknowledgements}
\addcontentsline{toc}{section}{Acknowledgements}

The authors thank Danica Ommen for pointing us to the inkjet application
and Patrick Buzzinni for providing us access to the inkjet data.
Additionally, we thank Philip Kegelmeyer for a careful reading of the
manuscript and insightful feedback.

\section*{Funding}\label{funding}
\addcontentsline{toc}{section}{Funding}

Sandia National Laboratories is a multimission laboratory managed and
operated by National Technology \& Engineering Solutions of Sandia, LLC,
a wholly owned subsidiary of Honeywell International Inc., for the U.S.
Department of Energy's National Nuclear Security Administration under
contract DE-NA0003525. This paper describes objective technical results
and analysis. Any subjective views or opinions that might be expressed
in the paper do not necessarily represent the views of the U.S.
Department of Energy or the United States Government. SAND2025-00008O.

\bibliography{references}

\newpage

\begin{center}
{\Large \textbf{Supplement}}
\end{center}

This document contains additional applications of the VEESA pipeline
corresponding to the article ``An Explainable Pipeline for Machine
Learning with Functional Data''. Section \ref{simS} compares the VEESA
pipeline to the cross-sectional approach with the simulated data.
Section \ref{hctS} contains additional details about the H-CT scan
example. Lastly, Section \ref{inkjetS} contains additional details about
the inkjet printer example.

\section{Comparing VEESA Pipeline to Cross-Sectional Approach with the
Simulated Data}\label{simS}

\begin{figure}

{\centering \includegraphics[width=3.5in]{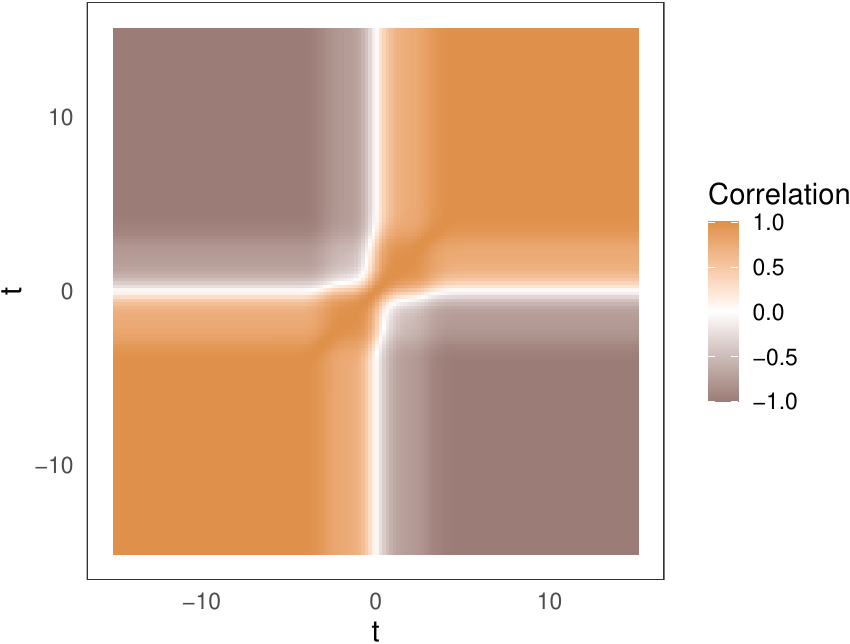} 

}

\caption{Spearman correlations between all pairs of cross-sectional variables from the simulated data.}\label{fig:figS1}
\end{figure}

The cross-sectional approach treats the observations across functions at
each time point as the predictor variables. An explainability method
such as PFI may then be applied to try to identify the times that are
important to the model for prediction. However, this approach presents a
disadvantage. Due to the nature of functional data, the cross-sectional
predictor variables are likely to be correlated. For example, consider
the simulated described in Section 2 of the main text. Figure
\ref{fig:figS1} shows a heatmap of all pairwise Spearman correlations
between the cross-sectional simulated data predictor variables. There
are strong positive and negative correlations for almost all variables.
As mentioned in Section 1 of the main text, correlation between
predictor variables leads to biased PFI results. Here, we apply the
cross-sectional modeling approach to the simulated data and compare the
results to the VEESA pipeline. We highlight the difficulties with
gaining insight to the model produced by the cross-sectional approach.

The simulated data is separated in the same training and testing
datasets with 400 and 100 observations, respectively. A random forest is
trained on the training data (\emph{randomForest} R package; version
4.7.1.1). The cross-sectional variables (one associated with each of the
150 times where the functions are observed) are treated as inputs. The
observed group is the response variable. The default tuning parameters
options are used to mimic the VEESA pipeline random forest from the main
text. The random forest performs well on the test data with an accuracy
of 1. The accuracy is slightly higher than the result from the VEESA
pipeline random forest (0.99). PFI is applied to the test data using a
metric of accuracy with 10 replications to mimic the computation of PFI
in the VEESA pipeline from the main text.

\begin{figure}

{\centering \includegraphics[width=5.5in]{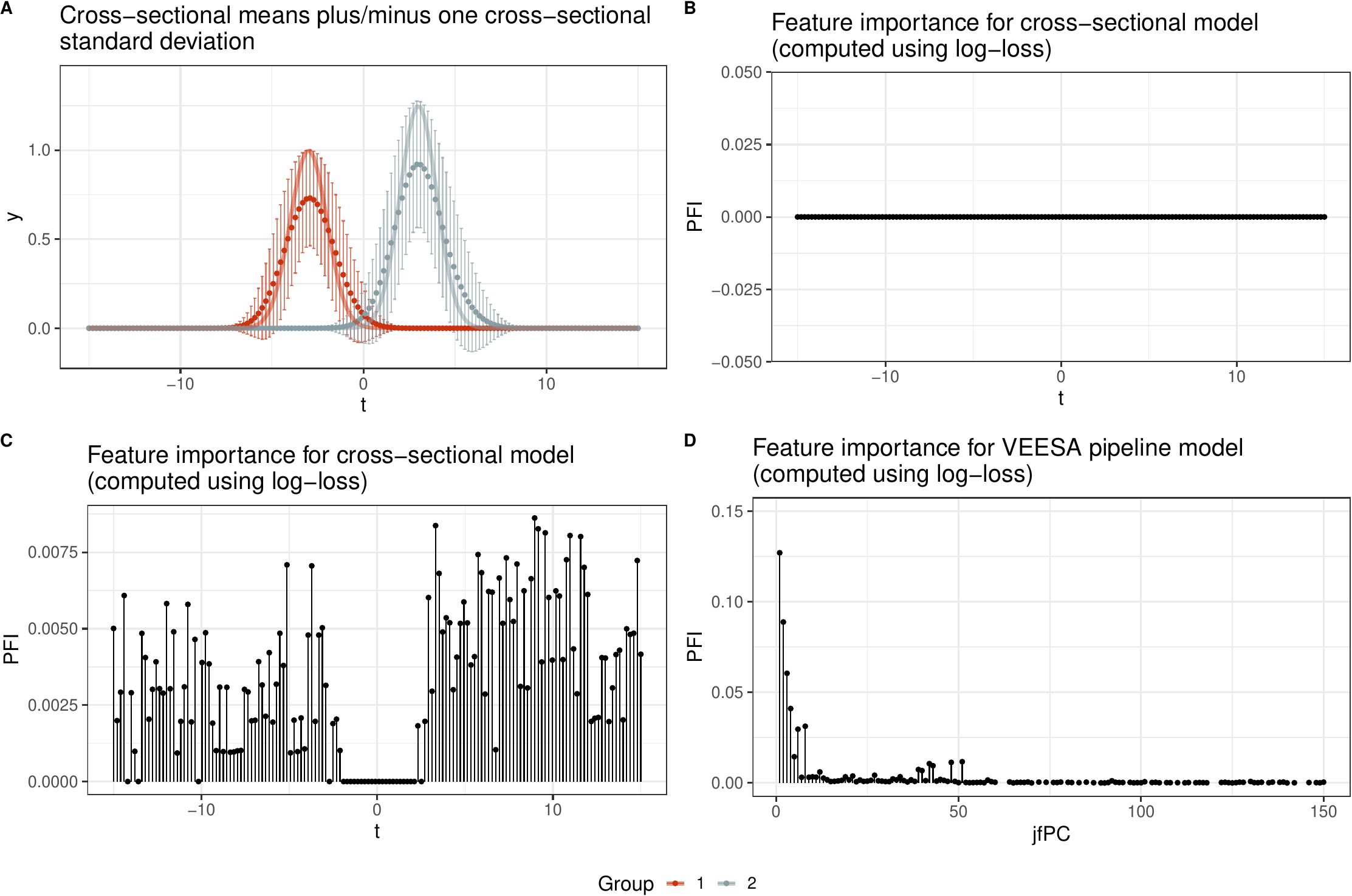} 

}

\caption{(A) The solid lines represent the true group means. The dots and error bars represent the cross-sectional group means plus and minus one cross-sectional standard deviation. (B-D) PFI values computed on the test data using the metric of accuracy for a random forest trained using the cross-sectional approach, using the metric of log-loss for a random forest trained using the cross-sectional approach, and the metric of log-loss for a random forest trained using the VEESA pipeline, respectively.}\label{fig:figS2}
\end{figure}

Figure \ref{fig:figS2}A includes the true group means depicted as solid
lines and the cross-sectional group means depicted by points. The error
bars represent plus/minus one cross-sectional standard deviation. As
seen in Section 2, the cross-sectional means do not capture the shape of
the true means. Figure \ref{fig:figS2}B depicts the PFI values computed
using the cross-sectional approach. Unlike the PFI results from the
VEESA pipeline method (Figure 3), the PFI values for all variables (time
points in this instance) are approximately 0 with no variability (all
times have standard deviations across PFI replicates of 0). In this
instance, the PFI results do not appear to be inflated due to bias.
Instead, this result suggests that none of the individual time points
are important in regards to the accuracy of the model. These PFI results
indicate that permuting one time point, while leaving all other time
points as observed, does not affect the model enough to alter the
accuracy of the model on the test data. This information is unhelpful in
identifying the aspects of the data that are important to the model for
prediction. The application of the VEESA pipeline to this data,
described in Section 3, produces non-zero PFI values, and the
interpretation of the most important fPC (jfPC 1) provides a clear
explanation as to the aspect of the data that is important to the model
for prediction.

Since the predictor variables in the cross-sectional approach are highly
correlated, we suspect that the non-zero PFI values are the result of
accuracy being used as the metric. Consider the definition of accuracy.
For a set of \(i=1,,,.n\) observations, let \(y_i\) be the observed
value and \(\hat{y}_i\) be the predicted value for observation \(i\).
Accuracy is computed as
\[Accuracy(y,\hat{y})=\frac{1}{n}\sum_{i=1}^nI(\hat{y}_i=y_i).\] With
the simulated data, the predicted value is whether a function belongs to
group 1 or 2. To compute the PFI for a time point, the observations
across the simulated functions at that time are permuted. With the
remaining 149 time points being highly correlated with the permuted
variable, enough information is likely provided to the model to produce
the same prediction as to which group the function belongs to. However,
a metric for a binary response variable based on the model probability
that a function belongs to a group is more likely to be affected by the
permutation of a single time point. One example of such a metric is
log-loss. For \(y\in\{0,1\}\) and \(p=P(y=1)\) (estimated from a model),
log-loss is defined as \[LL(y,p)=y\log(p)+(1-y)\log(1-p).\]

To test this idea, we compute new PFI values using the log-loss metric
for the random forest from the cross-sectional approach and the random
forest in the VEESA pipeline. Both sets of PFI values are computed on
the test data. Figures \ref{fig:figS2}C and \ref{fig:figS2}D depict the
PFI values computed for the cross-sectional and VEESA pipeline
approaches, respectively, using log-loss. As suspected, the log-loss
metric produces non-zero values. However, note the difference in y-axis
between the cross-sectional and VEESA pipeline PFI values. The
cross-sectional PFI values are much smaller than the VEESA pipeline PFI
values (i.e., the model predicted probability is much less affected when
a variable is permuted). The cross-sectional PFI values suggest that the
times between, approximately, \(t=-2\) and \(t=2\) have little
importance. The time points outside of this region have non-zero values.
We may expect the times between -2 and 2 to have little importance since
there is a lot of overlap between the two group during this time. We may
also expect the times in the ranges of \(t\in(-7,-2)\) and \(t\in(2,7)\)
to be important since the groups are distinguished by their differing
peaks associated with these intervals. However, the PFI values continue
to be non-zero in the below -7 and above 7, where the the functions all
have values of approximately zero. This result either suggests that the
model is doing a poor job of using the information from the predictor
variables, or more likely, these PFI values are biased due to the
correlation between predictor variables.

\begin{figure}

{\centering \includegraphics[width=5.5in]{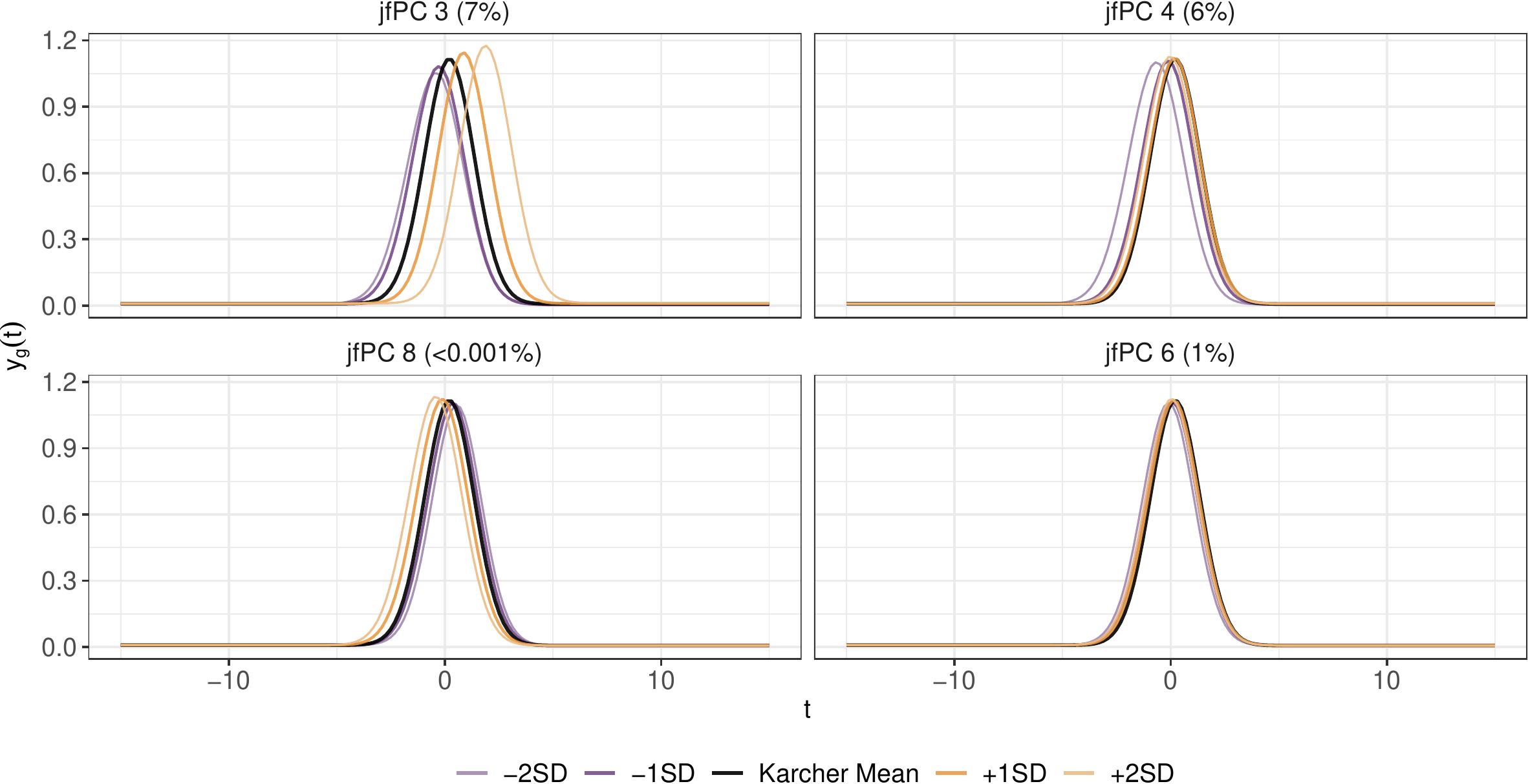} 

}

\caption{Principal direction plots of the VEESA pipeline jfPCs with log-loss based PFI values greater than 0.02 (excluding jfPCs 1 and 2 shown in Figures 2 and 4).}\label{fig:figS3}
\end{figure}

The VEESA pipeline PFI values computed with accuracy (Figure 3) only
found jfPCs 1 and 2 to be important for the test data. The log-loss
metric identifies additional principal components as important, but
jfPCs 1 and 2 remain the most important. Figures 2 and 4 showed that
jfPC 1 and jfPC 2, respectively, capture variability in both the
horizontal and vertical directions. Figure \ref{fig:figS3} includes the
principal direction plots for the four other principal components with
PFI (computed using log-loss) greater than 0.02. jfPC 3 focuses on
variability between functions with peaks similar to the mean around time
0 and functions with high peaks later than time 0. jfPCs 4 and 8 focus
mostly on the horizontal variability between functions. jfPC 6 focuses
on small amounts of horizontal variability. Although the VEESA pipeline
results in a smaller (but similar) test data accuracy than the
cross-sectional method, the feature importance results are more
meaningful. The results are understandable in the data space of the
problem and provide evidence for how the model is using the data for
prediction.

To summarize, there are several concerns with the cross-sectional
approach:

\begin{itemize}
\item
  The functional nature of the data is ignored. By treating each time
  point of samples as a predictor variable, the model is not aware of
  the relationship between samples within a function. By presenting the
  functions to the model in this form, information is lost. In some
  cases, the loss of information may lead to a loss in prediction
  performance.
\item
  The horizontal variability in the functions is ignored. In the
  simulated data, the true functional means have different peak times,
  which leads to horizontal variability in the functions. The
  cross-sectional approach only uses the variability in the functions in
  the vertical direction to discriminate between the two classes. This
  effect is seen in the cross-sectional group means in Figure
  \ref{fig:figS2}A. The shapes of the cross-sectional means do not
  accurately reflect the true functional means, because only the
  variability in the vertical direction is accounted when computing
  cross-sectional means in this manner.
\item
  Correlation between the predictor variables can lead to bias in PFI.
  In practice, it may not be clear which PFI values are biased, so it
  becomes difficult to trust the results.
\end{itemize}

All three of these concerns are addressed by the VEESA pipeline through
the use of efPCA.

\section{H-CT Material Classification Additional Analyses}\label{hctS}

Additional details about the analysis of the H-CT material
classification data are provided here. In particular, we present results
comparing model accuracies from the cross-sectional approach, the three
types of efPCA, and varying number of box-filter runs. We also include
the PFI variability from the implementations of the VEESA pipeline
included in the main text and the principal directions from the six most
important jfPCs identified in the main text.

\subsection{Smoothing Selection and Cross-Sectional
Comparison}\label{smoothing-selection-and-cross-sectional-comparison}

In order to select the number of box-filter iterations to use for the
VEESA pipeline analysis of the H-CT material classification data, we
consider a range of iterations and determine which leads to the best
accuracy result on the test data. In the process, we implement the VEESA
pipeline using jfPCA, vfPCA, and hfPCA, so we can compare their
predictive performances. We also implement several variations of the
cross-sectional approach for comparison. The details of the
implementations for smoothing, the VEESA pipeline, and the
cross-sectional approaches are described in the following paragraphs.

\emph{Smoothing} A box-filter method is applied to smooth the H-CT data.
The number of times the box-filter is run affects the smoothness of the
functions (i.e., more runs lead to smoother functions). In order to
determine the number times to run the box-filter, the box-filter is run
1, 5, 10, 15, 20, and 25 times.

\emph{VEESA Pipeline} The VEESA pipeline is applied to all versions of
the smoothed data using jfPCA, vfPCA, and hfPCA. A neural network is
used as the model for each scenario. All models are trained with the
default settings in \emph{scikit-learn} (one layer with 100 neurons).
The model accuracies are computed on the training and testing dataset.

\emph{Cross-Sectional Approach} The cross-sectional approach is applied
for comparison to the VEESA pipeline. Three variations in regards to
smoothing are considered: applying the cross-sectional approach with (1)
no smoothing, (2) after smoothing, and (3) after smoothing and ESA
alignment. Again, neural networks are used as the model (trained in
\emph{scikit-learn} with default settings).

\begin{figure}

{\centering \includegraphics[width=5.5in]{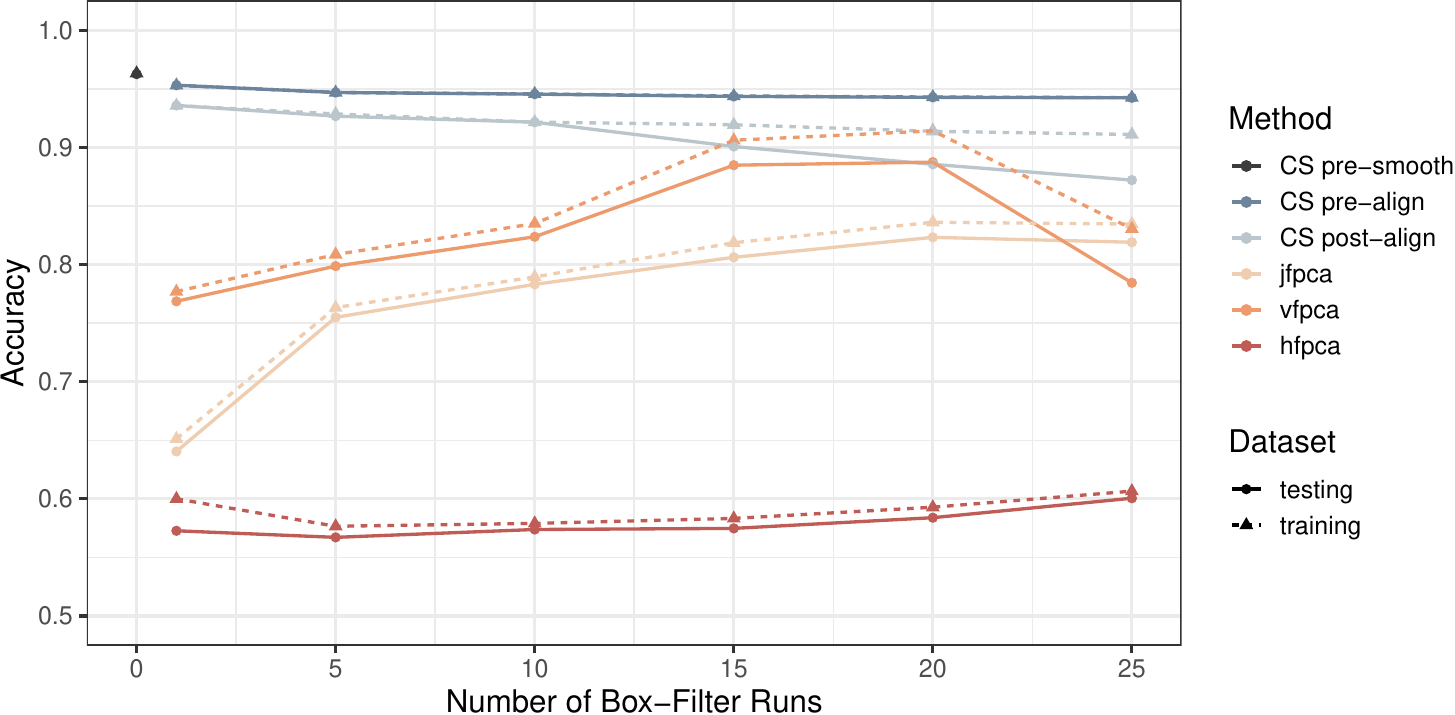} 

}

\caption{Model accuracies from neural networks applied using the VEESA pipeline and the cross-sectional approach. }\label{fig:figS4}
\end{figure}

The model accuracies versus the number of box-filter runs are depicted
in Figure \ref{fig:figS4} for all implementations. The colors represent
the method used to process the data before training a neural network.
The solid lines and circles indicate accuracy on the training data, and
the dashed lines and triangles indicate accuracy on the testing data.
The application of the cross-sectional method with no smoothing or
alignment returns the highest accuracy. This is followed by the
cross-sectional method with smoothing but no alignment with relatively
consistent accuracy regardless of number of times the box-filter is run.
The accuracies returned from the VEESA pipeline approaches are always
lower on the testing data than training data. hfPCA results in the
lowest accuracies followed by jfPCA, and vfPCA returns the highest
accuracy for all box-filter iterations until 25. This is expected since
it is understood that the vertical variability in the signatures
contains the most information for material classification. The results
from the VEESA pipeline using vfPCA and jfPCA when 15 runs of the
box-filter are those displayed in the main text.

Even though the cross-sectional approaches lead to the largest
accuracies (close to or above 0.9), the VEESA pipeline using vfPCA
achieves accuracy values close to the cross-sectional accuracies when
the box-filter is applied 15 and 20 times. In a high-consequence
application, sacrificing some predictive performance for a method that
returns more trustworthy explanations may be a necessity. However, no
tuning of the neural network parameters was done, and work in this area
could lead to increased predictive performance with the VEESA pipeline.

These results suggests two ideas for future work. It would be beneficial
to compare the predictive performance of the VEESA pipeline to the
cross-sectional approach in applications where horizontal variability
contains predictive information to see if the explicit modeling of the
horizontal variability affects performance. This analysis only considers
training and testing sets. It would be beneficial to see how the
cross-sectional and VEESA pipeline approach perform on a held out
validation dataset.

\subsection{PFI Variability and jfPCA Principal
Directions}\label{pfi-variability-and-jfpca-principal-directions}

\begin{figure}

{\centering \includegraphics[width=5.5in]{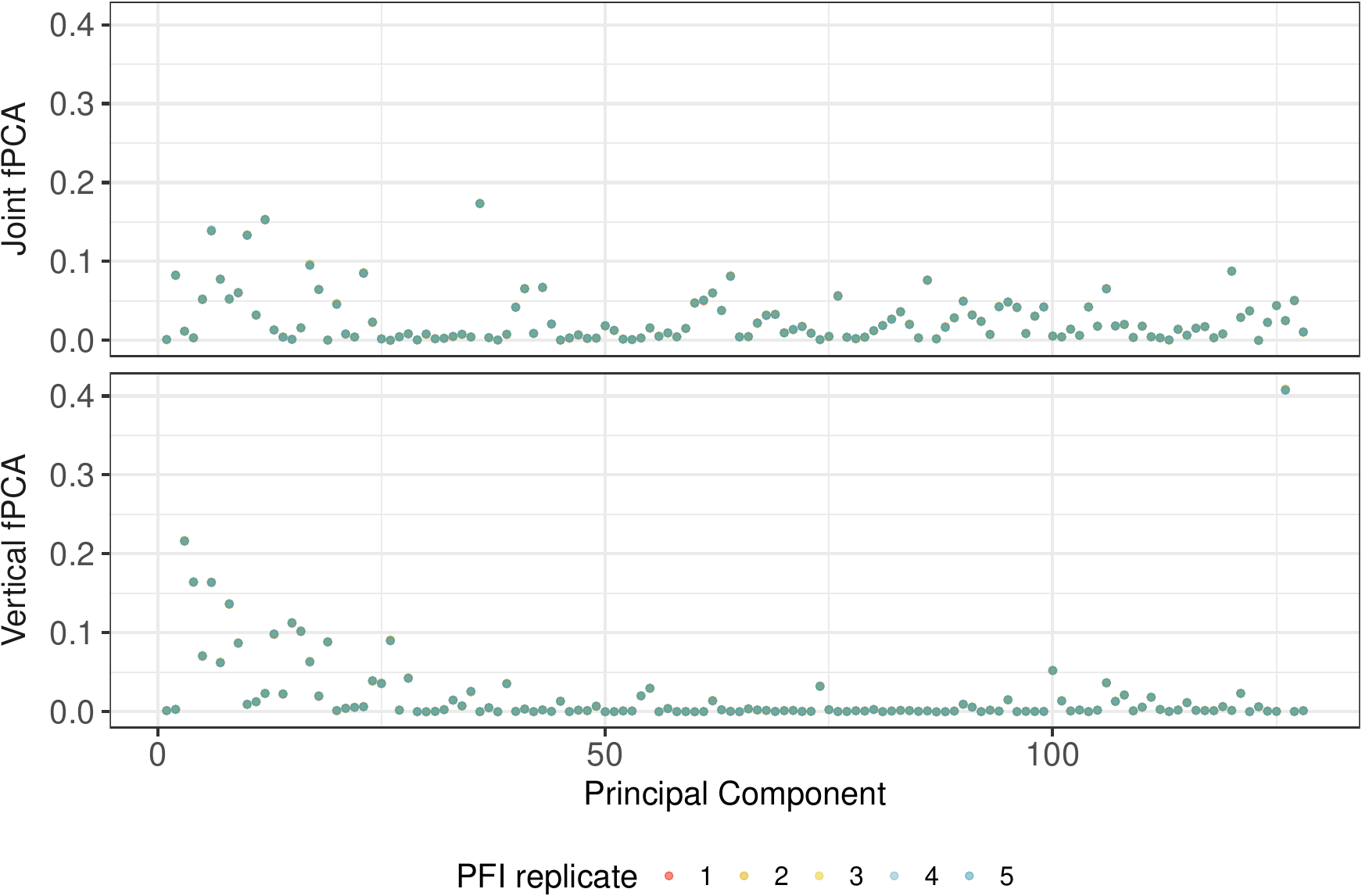} 

}

\caption{PFI replicate values for jfPCA (top) and vfPCA (bottom) associated with the H-CT material classification example.}\label{fig:figS5}
\end{figure}

Figure \ref{fig:figS5} depicts the five feature importance replicates
used to compute the PFI results included in the main text. The points
are colored by the replicate with an alpha shading that provides
transparency. However, all points appear to be the same color, due to
them overlapping. There is minimal amount of variability between the
feature importance replicates.

\begin{figure}

{\centering \includegraphics[width=5.5in]{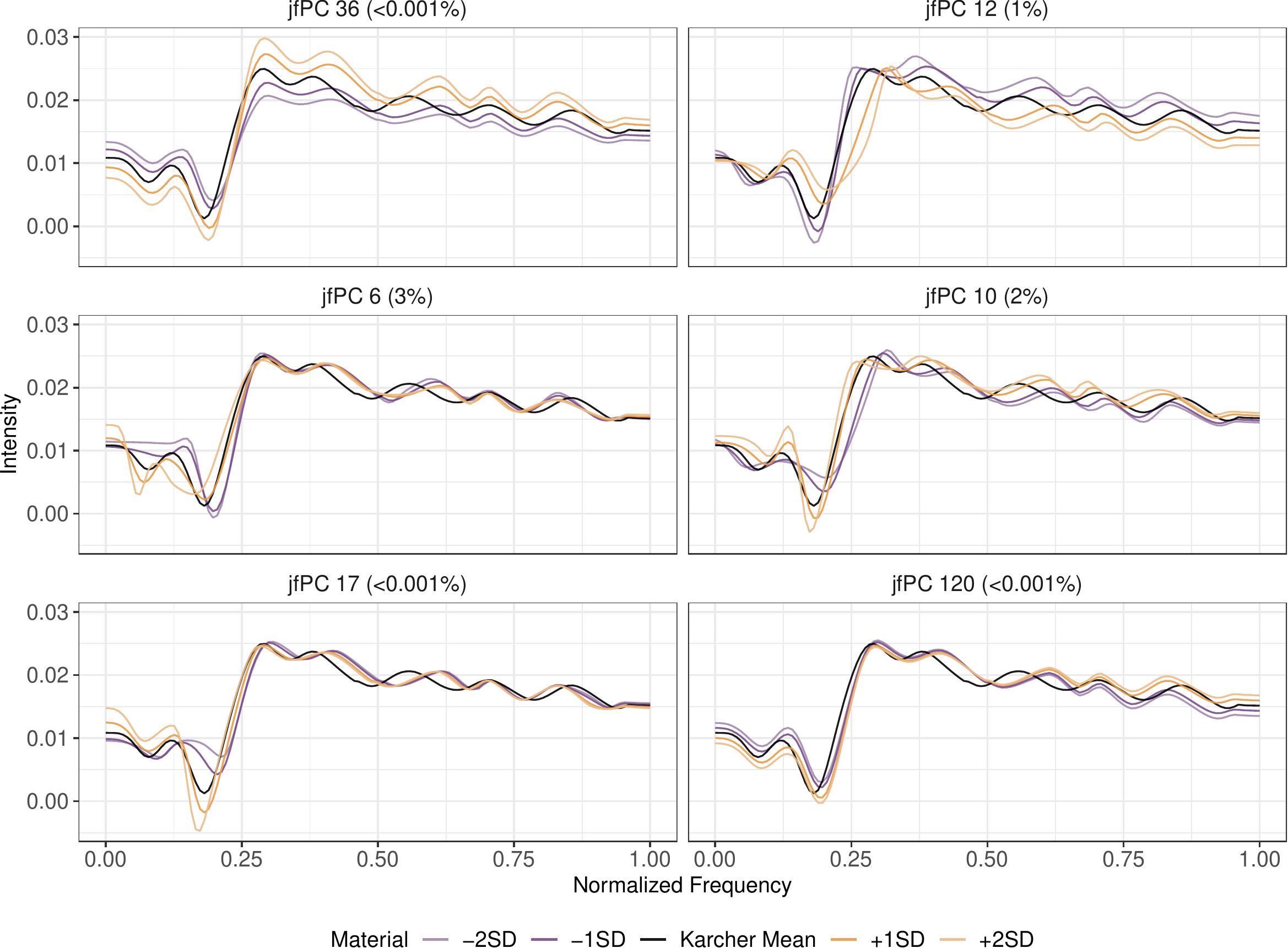} 

}

\caption{Principal directions from the six jfPCs with the highest PFI from the H-CT material classification example.}\label{fig:figS6}
\end{figure}

Figure \ref{fig:figS6} depicts the principal direction plots of the four
PCs with the highest PFI values from the jfPCA model. All four jfPCs
capture aspects of both horizontal and vertical variability. The PC with
the highest PFI (jfPC 36) focuses on vertical variability between
functions with a horizontal shift from the mean at frequencies larger
than 0.25. jfPC 36 also captures the change in variability around a
frequency of 0.25.

\section{Inkjet Printer Additional Analyses}\label{inkjetS}

We include some additional results from the inkjet printer analyses
here: classification performance within each printer, principal
directions for importance fPCs with magenta and yellow signatures, and
an implementation of the VEESA pipeline for the worst performing
scenarios with a subset of the principal components selected based on
the feature importance in the main text.

\subsection{Printer Classification
Details}\label{printer-classification-details}

\begin{figure}

{\centering \includegraphics[width=5.25in]{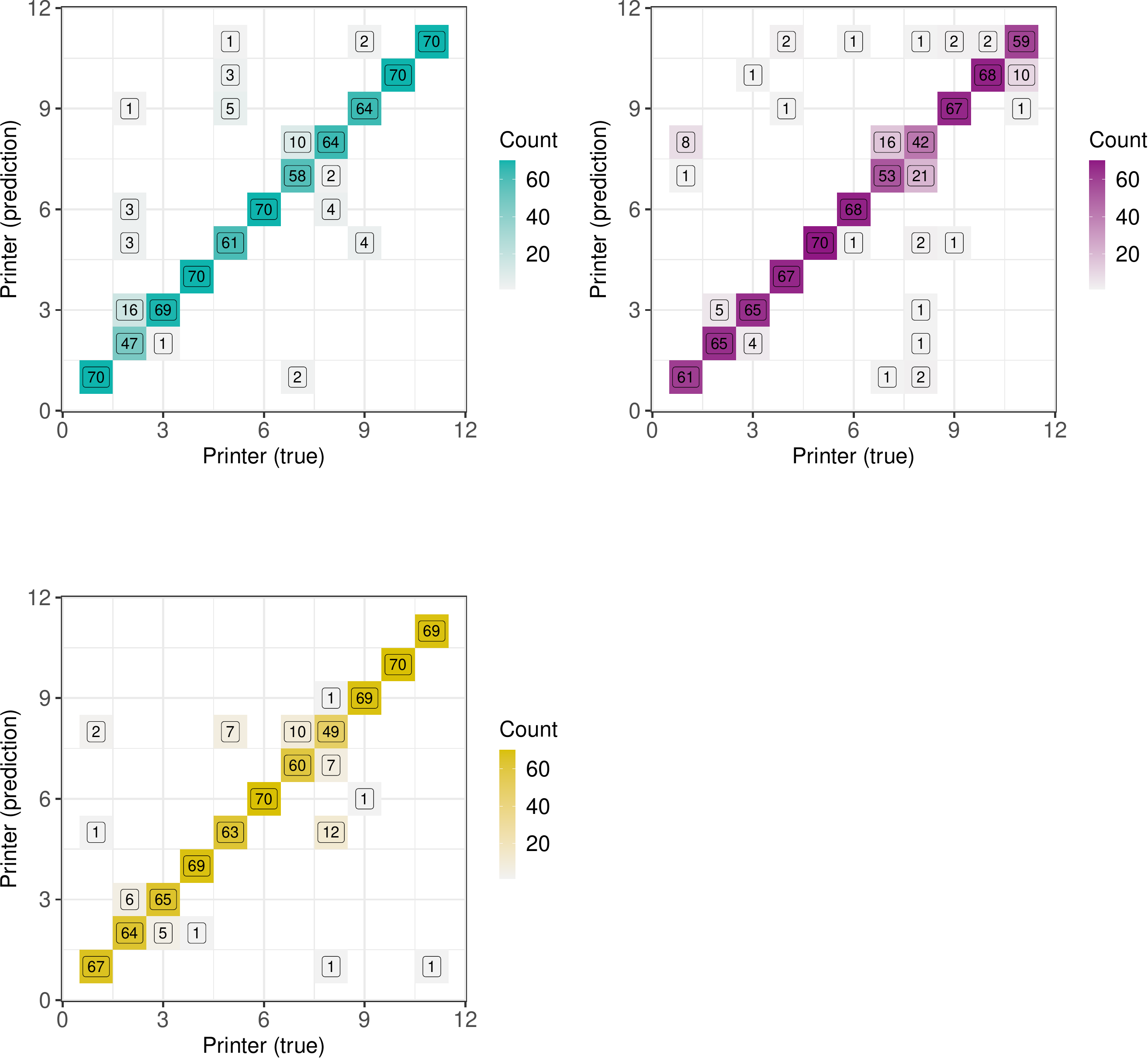} 

}

\caption{Confusion matrices for the inkjet printer random forests.}\label{fig:figS7}
\end{figure}

Figure \ref{fig:figS7} depicts confusion matrices for the three colors
considered in the inkjet printer analysis in the main text. The true
printer is listed on the x-axis, and the predicted printer is listed on
the y-axis. The plots are generated using all test fold predictions for
the best performing model for each color. These figures let us see which
printers are challenging. For all colors, printers 7 and 8 often are
predicted as the other printer. This is particularly true for magenta.
Printers 7 and 8 have the same manufacturer and model, so it is
understandable why these printers are more challenging. With cyan,
printer 2 also stands out as being predicted incorrectly as printer 3.
These two printers share the same manufacturer. With magenta, printer 11
is frequently mistaken for printer 10. Again, these printers share a
manufacturer. With yellow printers 2 and 3 also are incorrectly
predicted as the other printer. These results that the models tend to
struggle to distinguish printers that are from the same manufacturer,
which would be expected.

\subsection{Principal Directions from Magenta and Yellow
Signatures}\label{principal-directions-from-magenta-and-yellow-signatures}

\begin{figure}

{\centering \includegraphics[width=5.5in]{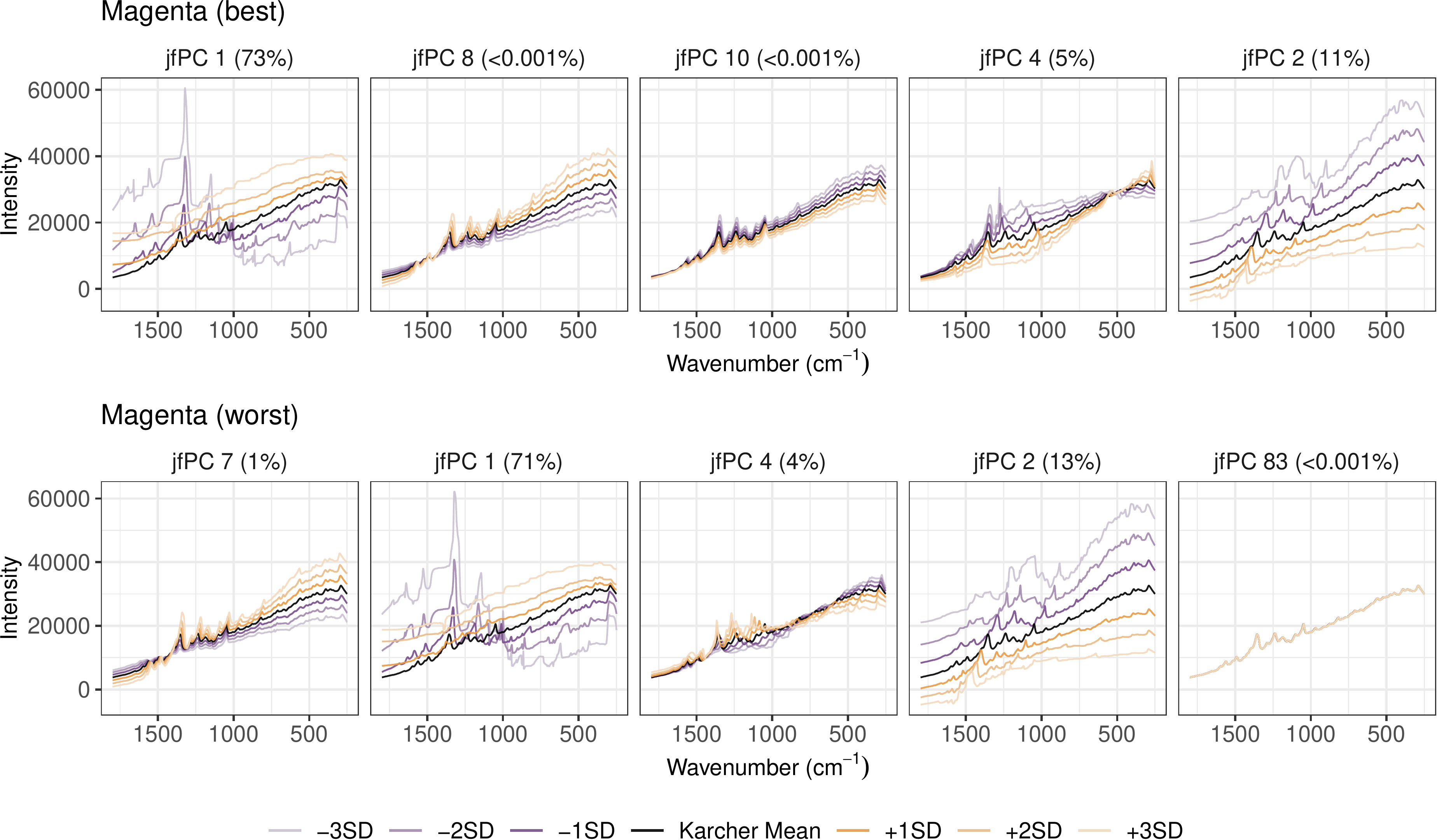} 

}

\caption{Principal directions from the best (top row) and worst (bottom row) models for predictions with magenta inkjet signatures. The jfPCs selected are those with the largest PFI values for their respective model. PCs are ordered from left to right based on highest to lowest feature importance.}\label{fig:figS8}
\end{figure}

Figure \ref{fig:figS8} shows the six most important PCs for the best
(top) and worst (bottom) magenta models. The most important jfPCs for
the best performing magenta model capture more horizontal variability
than the jfPCs for cyan. For example, the most important jfPC for
magenta (jfPC 1) captures variability between signatures that have peaks
occurring immediately before and after 1250 cm\(^{-1}\) and signatures
that are smoother besides for a slight increase around a wavenumber of
1200 cm\(^{-1}\). This first PC appears to capture the variability in
the magenta signatures between some printers that have dramatic peaks
around 1250 cm\(^{-1}\) (e.g., printer 7) and other printers that
gradually increase without such peaks and then decrease after 500
cm\(^{-1}\) (e.g., printer 4). The important fPCs for the worst magenta
model capture similar aspects of the functional variability but with
noisier functions.

\begin{figure}

{\centering \includegraphics[width=5.5in]{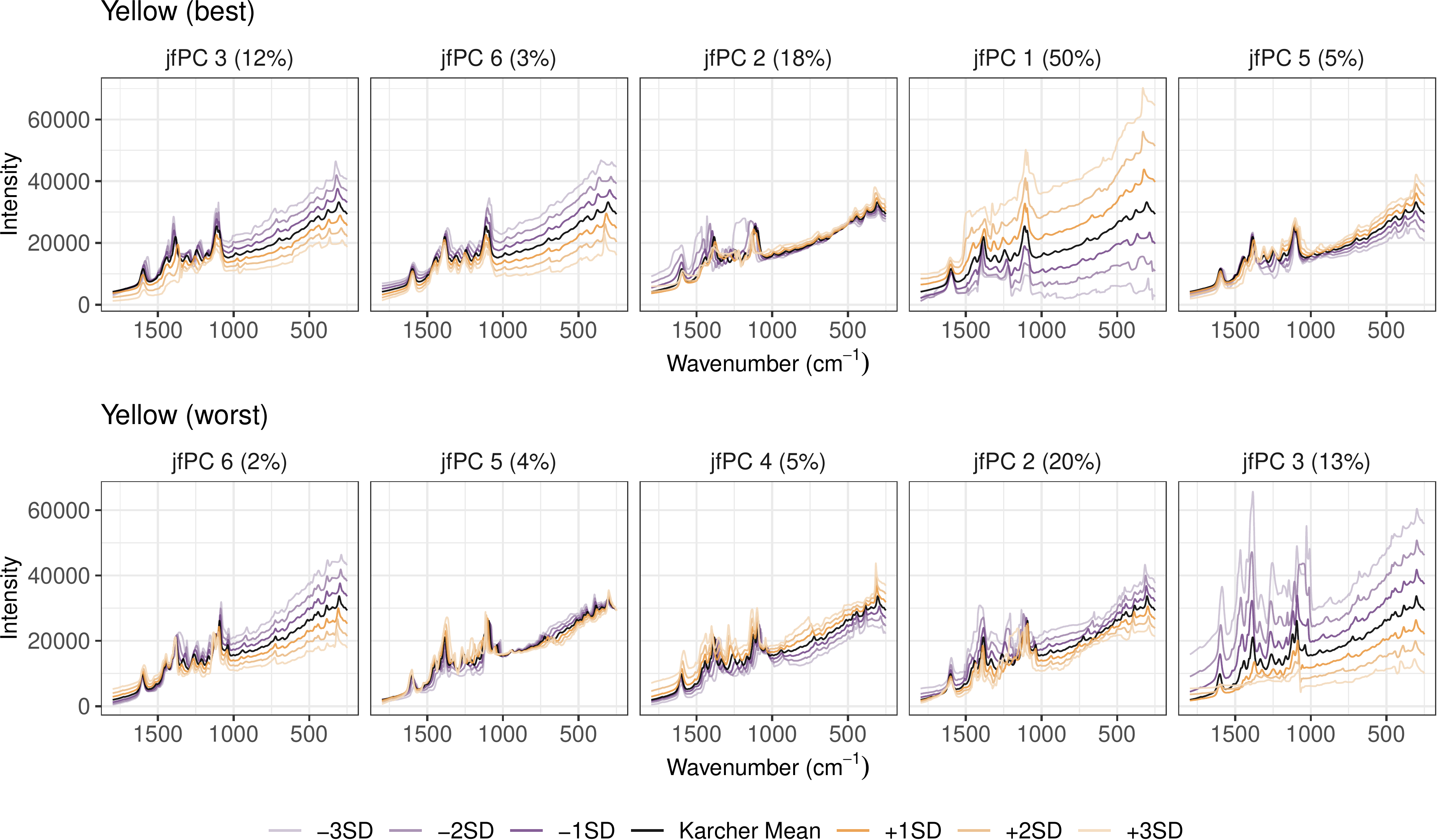} 

}

\caption{Principal directions from the best (top row) and worst (bottom row) models for predictions with yellow inkjet signatures. The jfPCs selected are those with the largest PFI values for their respective model. PCs are ordered from left to right based on highest to lowest feature importance.}\label{fig:figS9}
\end{figure}

Figure \ref{fig:figS9} shows the six most important PCs for the best
(top) and worst (bottom) yellow models. The most important jfPCs for the
yellow signature predictive model capture a combination of vertical
variability across all wavenumbers and the horizontal variability around
the peaks between wavenumbers of 1500 and 1000 cm\(^{-1}\). The yellow
signatures for all printers in Figure 9 have peaks in the range of
wavenumbers of 1500 and 1000 cm\(^{-1}\) (unlike the magenta
signatures), and there is more variability in the peaks in this range
for the yellow signatures compared to the cyan signature peaks. Similar
to cyan and magenta signatures, the amount that the signatures tends to
increase as the wavenumber increases varies by printers, which is picked
up by the important principal components. jfPC 2 for the yellow
signatures is particularly interesting in that it appears to capture
variability between signatures that have lots of peaks or noisy peaks in
the range of 1500 and 1000 cm\(^{-1}\) (e.g., Printer 4) compared to
signatures that do not have as many peaks (e.g., Printer 1). Again, the
fPCs for the worst yellow model capture similar aspects of the
variability but with noisier functions.

\subsection{Improving Models Using
PFI}\label{improving-models-using-pfi}

\begin{figure}

{\centering \includegraphics[width=5.5in]{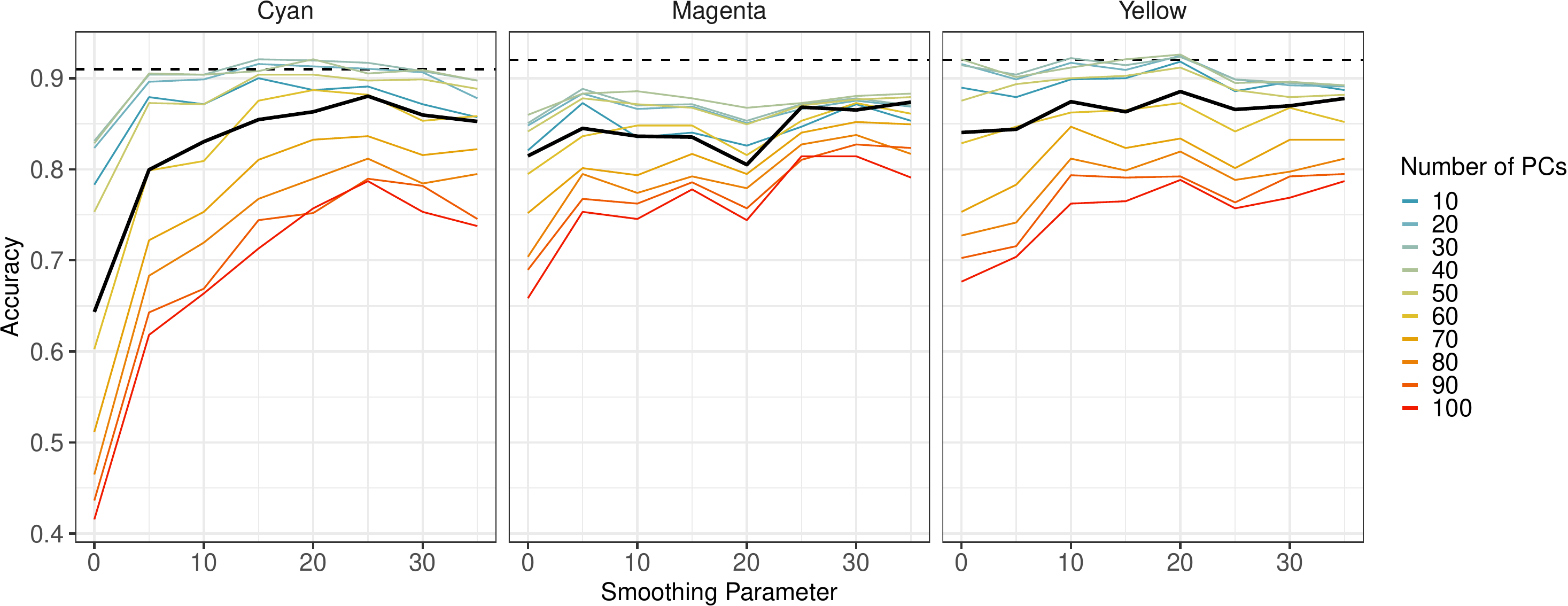} 

}

\caption{Cross validation average accuracies. Black lines represent CV accuracies from models with PC selected via PFI.}\label{fig:figS10}
\end{figure}

As a last step in our analysis, we are interested in determining if we
can make use of the information from the PFI results of the worst models
to improve the predictive performance. We do this by implementing the
cross validation procedure describe in the main text. However, for this
implementation, we only consider random forests with 500 trees. We
choose 500 trees, because it falls between the number of trees
associated with the best performing models (i.e., 250 or 1000 trees).
Additionally, we use PCs 1 to 25 and PCs 75 to 100 for all models based
on the areas with important PCs as seen in Figure 11. We consider the
same range of smoothing iterations (0, 5, 10, 15, 20, 25, 30, 35), and
again, random forests are implemented separately by color. Figure
\ref{fig:figS10} shows the average cross validation accuracies from this
analysis.

In Figure \ref{fig:figS10}, the plots of results are separated by color.
The number of smoothing iterations is depicted on the x-axis, and the
average accuracy is on the y-axis. The colored lines represent the
accuracies from our original analysis (Figure 10), and the color of the
lines indicates the number of PCs included in the model. The black line
represents the average cross-validation accuracy from the updated models
with the subset of fPCs selected based on the PFI results. For all
smoothing iterations, the down selection of fPCs based on PFI led to a
clear increase in test fold average accuracy. The accuracies are still
lower than the best performing models, but this provides a case study
where PFI provides useful information for improving model predictive
performance.

\end{document}